\documentclass[sn-mathphys-num]{sn-jnl}

\usepackage{graphicx}%
\usepackage{multirow}%
\usepackage{amsmath,amssymb,amsfonts}%
\usepackage{amsthm}%
\usepackage{mathrsfs}%
\usepackage[title]{appendix}%
\usepackage{xcolor}%
\usepackage{textcomp}%
\usepackage{manyfoot}%
\usepackage{booktabs}%
\usepackage{algorithm}%
\usepackage{algorithmicx}%
\usepackage{algpseudocode}%
\usepackage{listings}%
\usepackage{tabularx}

\theoremstyle{thmstyleone}%
\theoremstyle{thmstyletwo}%

\theoremstyle{thmstylethree}%
\usepackage{lineno}
\begin{document}



\title[Article Title]{Supporting renewable energy planning and operation with data-driven high-resolution ensemble weather forecast}


\author[1]{\fnm{Jingnan} \sur{Wang}}

\author[2]{\fnm{Jie} \sur{Chao}}

\author[3]{\fnm{Shangshang} \sur{Yang}}


\author[1]{\fnm{Kaijun} \sur{Ren}}

\author[1]{\fnm{Kefeng} \sur{Deng}}

\author[2]{\fnm{Xi} \sur{Chen}}

\author[4]{\fnm{Yaxin} \sur{Liu}}

\author[4]{\fnm{Hanqiuzi} \sur{Wen}}

\author[2]{\fnm{Ziniu} \sur{Xiao}}

\author[1]{\fnm{Lifeng} \sur{Zhang}}

\author[5]{\fnm{Xiaodong} \sur{Wang}}

\author*[1]{\fnm{Jiping} \sur{Guan}}\email{guanjiping@nudt.edu.cn}

\author*[2]{\fnm{Baoxiang} \sur{Pan}}\email{panbaoxiang@lasg.iap.ac.cn}

\affil[1]{\orgdiv{College of Meteorology and Oceanography}, \orgname{National University of Defense Technology}, \orgaddress{\city{Changsha}, \country{China}}}

\affil*[2]{\orgdiv{Institute of Atmospheric Physics}, \orgname{Chinese Academy of Sciences}, \orgaddress{\city{Beijing}, \country{China}}}

\affil[3]{\orgdiv{Earth System Modeling, School of Engineering and Design}, \orgname{Technical
University of Munich}, \orgaddress{\city{Munich}, \country{Germany}}}

\affil[4]{\orgname{China Huaneng Clean Energy Research Institute}, \orgaddress{\city{Beijing}, \country{China}}}

\affil[5]{\orgdiv{College of Computer}, \orgname{National University of Defense Technology}, \orgaddress{\city{Changsha}, \country{China}}}


\abstract{
The planning and operation of renewable energy, especially wind
power, depend crucially on accurate, timely, and high-resolution weather information. 
Coarse-grid global numerical weather forecasts are typically downscaled to meet these requirements, introducing challenges of scale inconsistency, process representation error, computation cost, and 
entanglement of distinct uncertainty sources from chaoticity, model bias, and large-scale forcing.
We address these challenges by learning the climatological distribution of a target wind farm using its high-resolution numerical weather simulations. 
An optimal combination of this learned high-resolution climatological prior with coarse-grid large scale forecasts yields highly accurate, fine-grained, full-variable, 
large ensemble of weather pattern forecasts.
Using observed meteorological records and wind turbine power outputs as references, the proposed methodology verifies advantageously compared to existing numerical/statistical forecasting-downscaling pipelines, regarding either deterministic/probabilistic skills or economic gains. 
Moreover, a 100-member, 10-day forecast with spatial
resolution of 1~km and output frequency of 15~min  
takes $<$1 hour on a moderate-end GPU, as contrast to 
$\mathcal{O}(10^3)$
CPU hours for conventional numerical simulation. 
By drastically reducing computational costs while maintaining accuracy, our method paves the way for more efficient and reliable renewable energy planning and operation. 
}




\maketitle

\section{Introduction}\label{sec1}
To tackle energy crisis and climate change, renewable energy production, including mainly wind and solar power, is steadily growing, taking up 30-35\% of global electricity generation by 2023 \cite{wiatros2024global}. 
Due to the weather-dependent nature of these resources,  
accurate, timely, and high-resolution weather information becomes instrumental for assessing and planning power plants, optimizing operations, mitigating risks and ensuring reliable energy production \cite{halloran2024data, agupugo2024optimization, xu2024resilience, damiani2024exploring}. 

Despite significant advancements in weather prediction techniques, current weather forecasting services fail to fully meet the needs of the renewable energy industry \cite{Buster2024HighresolutionMW}.
Operational forecasting centers routinely update their 0-10 day predictions at 10-50 km resolutions.
These coarse-grid predictions are thereafter finely resolved for individual wind farm and photovoltaic plant  \cite{craig2022overcoming}, so as to support power output prediction,  maintenance scheduling, and risk preparation.

This forecasting-downscaling pipeline faces multiple challenges.
First, we must carefully reconcile coarse-grid forecasts to initiate and bound the downscaling model.
Initial state estimates from crude interpolation requires
spinning-up to achieve spatiotemporal and inter-variable consistency. This spinning-up process may drift model's state away from large-scale constraints \cite{laprise2008regional,jerez2020spin}, and delay the delivery of high-resolution forecasts by up to 12-24 hours \cite{short2022reducing}. 
Moreover, imbalance at boundaries is often inevitable, as we attempt to preserve large-scale features from the global forecasting model, while allowing small-scale processes to develop locally. 
This imbalance can potentially introduce exponentially amplifying errors, 
precluding mid-range (i.e., beyond 3 days) simulations \cite{lo2008assessment}.
Another concern comes from the process representation error in the downscaling model. Many key meteorological processes that directly 
concern renewable energy providers
remain unresolved in the high-resolution model, such as radiation and turbulent transfer of momentum and heat \cite{stevens2019dyamond}. These unresolved processes set an upper limit to the accuracy obtainable by dynamical methods of
forecasting, beyond which we shall resort to 
empirical parameterizations that come with error-prone functional forms and free parameters \cite{stensrud2007parameterization}.
Last but not least, the computational cost is often prohibitively high to finely resolve the geophysical fluid dynamics \cite{wedi2014increasing,dujardin2022wind},  calibrate the parameterization schemes for accurate approximation of unresolved physical processes \cite{miyamoto2013deep,rasp2018deep}, or generate large forecast ensembles to provide uncertainty-aware guidance for end-users \cite{leutbecher2008ensemble,zhu2002economic}.

In spite of these operational challenges, high-resolution numerical simulations are arguably the most reliable approach for revealing detailed weather patterns that emerge from the interplay of physical laws, topography, and cyclical phenomena \cite{satoh2019global, stevens2019dyamond}. If we could identify and exploit these regularities, we may sidestep the need for exhaustive geophysical computations, and avoid many of the associated difficulties. 
This insight has empowered machine learning techniques for efficient and accurate predictions \cite{buster2024high}.
Nevertheless, all learning algorithms inherently embed inductive biases about the problem domain in their architecture design or objective function choice \cite{bronstein2017geometric}, which fundamentally shape how the algorithms model and interpret the training data.
Specifically, inductive biases for model architecture design, such as smoothness (used in nearest neighbor methods) or spatial locality and translation invariance (employed in convolutional neural networks), can enhance model generalization, efficiency, and interpretability \cite{pan2019improving}. However, these same biases may significantly constrain model performance when the underlying data patterns don't align  well with the assumed structure \cite{PhysRevResearch.5.043252}.
Moreover, downscaling models trained in a supervised manner often 
aim at minimizing squared error loss averaged over different variables, regions, and forcing conditions. This objective function implicitly assumes stationary, independent, normally distributed fitting errors, resulting in over-smoothing outcomes that underestimate extremes \cite{pan2021learning}. 
To avoid these pitfalls, machine learning downscaling models must find effective priors that capture the regularities and complexities in high resolution meteorological data \cite{bronstein2017geometric}.

Typically, these domain-specific, highly-informative priors are not readily available, as there is no shortcut solution to determine regularities unfolding along chaotic evolution of a geophysical fluid dynamics system \cite{wolfram2002computational}. 
Yet, feasible solutions are within reach.  
Recently, deep generative models, in particular, probabilistic diffusion models \cite{ho2020denoising,song2020score,song2021solving}, have established themselves as powerful tools for learning expressive priors from data \cite{feng2023score,bouman2023generative,habring2024neural}.
Here, we present a diffusion-based weather research and forecasting (DWRF) framework, which deploys a probabilistic diffusion model to learn climatological prior distribution for a target wind farm, using
its high-resolution numerical weather simulations as training data. 
This learned climatological prior captures intricate inter-variable and spatial regularities within the finely resolved meteorological field data \cite{page2021revealing}. 
We thereafter combine this learned climatological prior with coarse-grid large scale
forecasts in a \textit{plug-and-play} manner \cite{6737048}, so as to minimize ensemble prediction error \cite{pan2023probabilistic}, hence generating highly accurate, fine-grained, full-variable, uncertainty-aware forecasts.




We compare the proposed methodology with classical numerical models and other deterministic/probabilistic machine learning approaches, benchmarked against observed meteorological records and wind turbine power output data. Particularly, we focus on addressing the scale consistency, computation efficiency, and uncertainty quantification challenges that hinders existing approaches. We demonstrate the unique advantage of the proposed methodology regarding its accuracy, efficiency, adaptability, and uncertainty awareness, highlighting its immediate applications in renewable energy planning and operations. 

\section{Diffusion-based high-resolution weather forecasting}

To overcome the forecasting limitations described earlier, our approach begins with capturing the fundamental structure of regional atmospheric behavior. Weather patterns exhibit complex dynamics that manifest across multiple spatial and temporal scales with intricate interdependency. We characterize these patterns by modeling the probabilistic distribution of atmospheric states using a diffusion-based framework trained on detailed climate simulations. This statistical representation encapsulates the relationships between meteorological variables throughout our domain, identifying the underlying manifold where realistic weather configurations reside. It further quantifies the probability distribution of possible atmospheric states, providing critical insight into how weather systems naturally evolve within our region of interest.

\begin{figure}[h]
\centering
\includegraphics[width=0.9\textwidth]{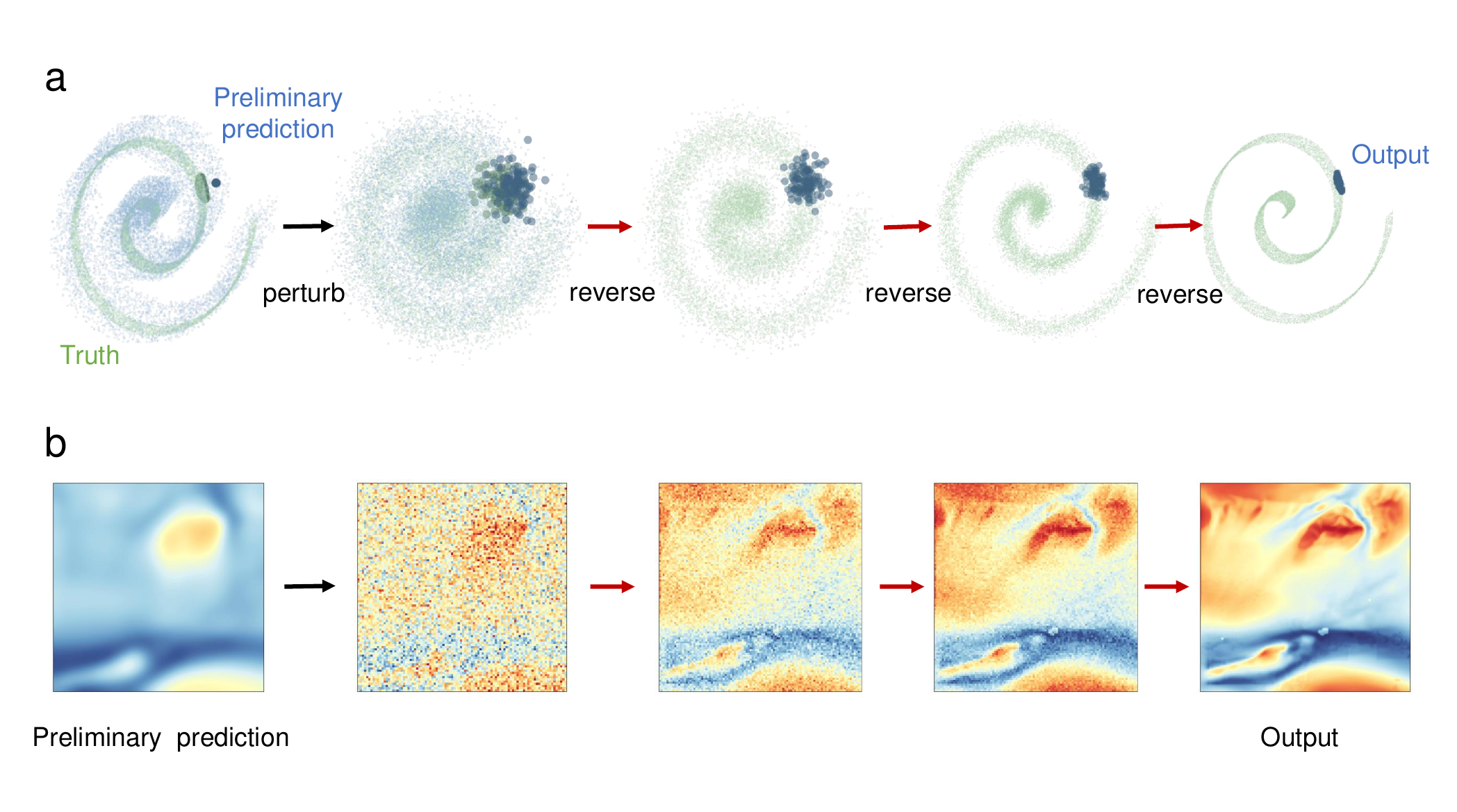}
\caption{Schematic illustration of guiding an unconditional diffusion model in DWRF. (a) The green dots and contours represent the truth and its distribution, while the blue ones represent the preliminary prediction and its distribution. The distribution of the ground truth can be obtained from an unconditional diffusion model to learn its climatology through a self-supervised learning manner (see Sec.\ref{method-diffusion}). The preliminary prediction comes from a regression model that turns low-resolution input into high-resolution output in a supervised manner. The leftmost image shows the initial state where the preliminary prediction mismatches the truth. Given the preliminary prediction, we first perturb it with Gaussian noise and followed by the reverse Stochastic Differential Equation (SDE) process of the diffusion model to progressively remove the noise and refine the prediction towards the ground truth distribution. (b) A concrete example of the 2m-temperature prediction process. Starting from a smooth preliminary prediction (leftmost), the image undergoes perturbation (becoming noisy), followed by several reverse steps that gradually refine and sharpen the features, resulting in a detailed final output (rightmost).
 }\label{fig1}
\end{figure}

DWRF leverages this learned probabilistic representation as a sophisticated prior that serves multiple functions: it guides the refinement of coarse-resolution forecasts, ensures physical consistency across meteorological fields, and generates realistic variability reflecting unresolved sub-grid processes (Fig. 1). This is accomplished through a carefully calibrated noise addition and sequential denoising methodology. When given a preliminary prediction containing systematic biases from inadequate resolution or parametrization, DWRF first perturbs this prediction with noise calibrated to the confidence in the initial forecast. The subsequent reverse diffusion process progressively refines the forecast toward the learned distribution of physically plausible states. This creates an iterative refinement cycle that balances fidelity to large-scale forcing while incorporating the high-resolution patterns and physics learned from historical simulations.

\section{Results}\label{sec2}

\subsection{High-resolution simulation informed climatological prior}

We first evaluate how the unconditional diffusion model in DWRF reproduces the high-resolution simulation informed climatological prior distribution. To establish a physics-constrained climatological prior for renewable energy applications, we focus on a 100 km$\times$×100 km wind farm situated in China's Northwestern Gobi Desert – a region characterized by complex terrain-induced turbulence and frequent dust storms that challenge conventional forecasting systems. Leveraging the Weather Research and Forecasting (WRF) model \cite{skamarock2008description}, we perform dynamical downscaling of 25-km ERA5 climate reanalysis data \cite{hersbach2020era5} from 2021 to 2023 at 1-km horizontal resolution (see Methods). This high-resolution simulation explicitly resolves critical boundary-layer processes (e.g., momentum transfer and thermal convection) while preserving large-scale circulation patterns from global reanalysis, generating a 15-minute interval dataset.

Building upon this dynamically consistent dataset, we train a probabilistic diffusion model \cite{ho2020denoising} to approximate the joint climatological distribution of high-resolution meteorological variables. The probabilistic diffusion model is composed of a series of deep neural nets trained to denoise WRF simulation outputs corrupted to a hierarchy of noise levels. Once trained, these denoising neural nets are chained across noise levels to transform the target climatological distribution to and from certain structured, non-informative distributions \cite{chaoj2025,pan2025GAP}, i.e., isotropic Gaussian, enabling accurate probability distribution approximation, efficient sampling, and effective probabilistic inference. The learned climatological distribution can be seen as a prior distribution, which preserves several characteristics.

\begin{itemize}
\item Spatial coherence of terrain-forced flow patterns in high resolution
\item Inter-variable thermodynamic constraints 
\item Scale-aware turbulence spectra distributions
\end{itemize}

\begin{figure}[h]
\centering
\includegraphics[width=0.98\textwidth]{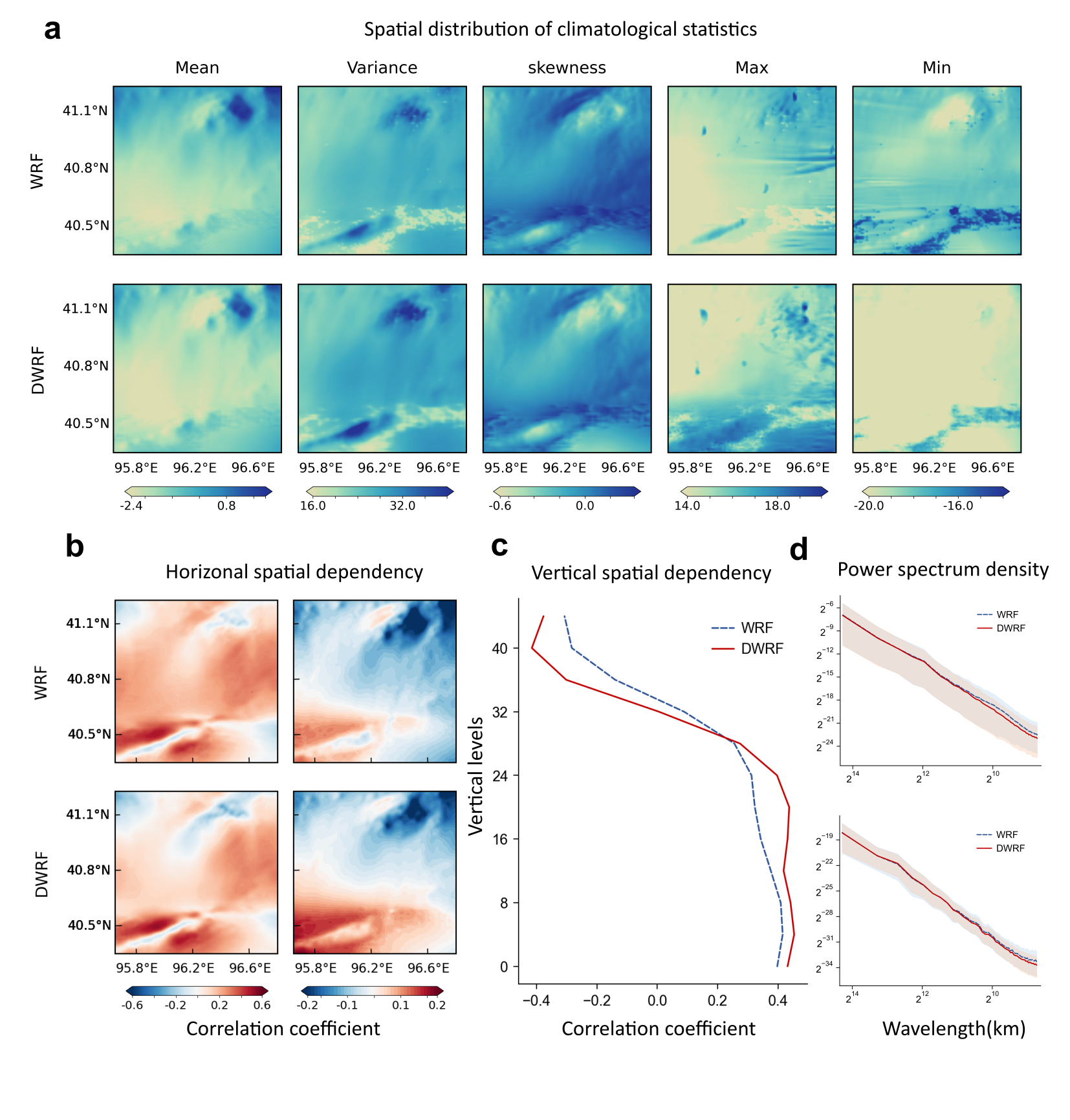}
\caption{Evaluation of the high-resolution climatological distribution derived from DWRF. (a) Spatial distributions of key climatological statistics (mean, variance, skewness, maximum, and minimum) from WRF and DWRF. The climatological statistics are both computed from 3000 random samples. (b) Horizontal spatial correlations between variables (u-v wind components on the left column and u-t on the right column). (c) Vertical correlation profiles across atmospheric layers between u wind components and temperature. (d) Power spectral density as function of wavelength for 10m wind (the top row) and 2m temperature (the bottom row).}\label{fig2-clima}
\end{figure}

To demonstrate this point, we compare the climatological distribution learned from the proposed DWRF with the WRF model, a mesoscale physics-driven NWP model. Fig. \ref{fig2-clima}a shows a comparison of statistical distributions (mean, variance, skewness, maximum and minimum) between WRF simulations and DWRF unconditional generations, given the 10m u wind component as an example.
DWRF demonstrates remarkable agreement with WRF simulations across key statistical moments of wind field distributions. For mean wind fields (the first column), DWRF not only faithfully replicates the synoptic-scale circulation patterns but also precisely resolves localized features such as the spatial positioning and intensity maxima of high-speed wind speed.
The spatial variance distribution (the second column), critical for assessing the temporal variability of wind resources in energy applications, reveals DWRF's capability to capture the location of turbulence intensity and their gradient transitions, enabling reliable quantification of short-term wind power fluctuations. 
For skewness patterns (the third column), the model successfully preserves the asymmetric characteristics of wind speed distributions. Such accuracy helps to evaluate the extreme event probabilities.
Moreover, DWRF exhibits superior performance in simulating extreme value distributions such as the maximum and minimum values (the fourth and fifth column). This advancement directly addresses the long-standing task of assessing the extreme wind event effects in wind power planning. These results show great consistency of spatial coherence between DWRF and WRF climatological distributions in high resolution.

Fig. \ref{fig2-clima}b presents the horizonal spatial correlation coefficients between u wind component and v wind component (the left column), u wind component and temperature (the right column). The pronounced consistency between DWRF and WRF in capturing inter-variable thermodynamic constraints underscores the success of DWRF’s AI-driven framework in internalizing the physically coherent relationships embedded in the NWP model. As demonstrated in Fig. \ref{fig2-clima}b, both models exhibit remarkably aligned horizontal spatial correlation coefficients for the u-v wind component pair. The vertical correlation profiles reinforce this consistency (Fig. \ref{fig2-clima}c). Both models show a characteristic decay of correlation magnitudes with height, mirroring the expected suppression of turbulent interactions in stable stratified layers.

Such alignment between the two models highlights that DWRF has learned not just statistical patterns but the dynamic consistency of the NWP system. The thermodynamic constraints between variables—such as the thermal wind balance linking u-wind and temperature gradients—are inherently nonlinear and sensitive to misrepresentation. This fidelity is particularly critical for operational applications, where preserving physical consistency across variables ensures robust predictions under diverse meteorological conditions.

Fig. \ref{fig2-clima}d compares the power spectral density (PSD) of DWRF and WRF simulations, illustrated for 10m wind (the top row) and 2m temperature (the bottom row). Power spectra are employed to evaluate the fidelity of two-dimensional spatial structures across scales. Unlike conventional AI models, which often exhibit spurious power decay at small scales due to incomplete physical constraints, DWRF resolves both large-scale atmospheric dynamics and fine-scale turbulent processes. Notably, DWRF accurately reproduces WRF’s spectral slopes—a hallmark of turbulence characteristics—indicating its capacity to capture not only climatological patterns but also the scale-dependent interactions governing high-resolution atmospheric processes. This spectral fidelity underscores DWRF’s success in internalizing the multi-scale physical constraints of the NWP system, ensuring dynamically consistent predictions across spatial resolutions. Such capability is pivotal for applications demanding precision in complex terrain, such as wind energy assessment, where resolving both large-scale forcings and microscale variability is essential.

This section establishes a physics-constrained climatological prior by integrating high-resolution numerical weather simulations with unconditional probabilistic diffusion models. Results collectively demonstrate that DWRF effectively learns physical constraints, producing dynamically coherent predictions essential for renewable energy applications in complex terrain. The model’s ability to replicate climatological distributions, thermodynamic relationships, and multi-scale spatial structures positions it as a reliable tool for high-resolution weather ensemble forecasting in renewable energy applications.

\subsection{Case studies}
Having established a physics-constrained climatological prior that captures the complex patterns and relationships within high-resolution meteorological fields, we now turn to its practical application in weather forecasting. Section 2.2 demonstrates how DWRF effectively combines this learned prior distribution with preliminary forecasts from a U-Net architecture to generate high-resolution ensemble predictions. The U-Net model provides an initial downscaled estimate based on coarse-resolution inputs, which serves as a starting point anchored to the large-scale atmospheric conditions. This preliminary forecast is then refined through the diffusion process guided by our climatological prior, striking an optimal balance between large-scale constraints and locally-consistent high-resolution features. By sampling multiple instances from this process with varied noise injection levels, DWRF generates a physically consistent ensemble that quantifies forecast uncertainty while preserving the detailed spatial patterns crucial for applications like wind energy prediction. The case studies that follow illustrate this approach's effectiveness in producing accurate, high-resolution ensemble forecasts that outperform traditional and alternative machine learning methods.

In this subsection, we evaluate the effectiveness of our proposed DWRF model through a comprehensive case study, focusing on two key meteorological variables: wind speed and temperature. We compare DWRF's performance against three alternative models representing different approaches to weather prediction: the Weather Research and Forecasting (WRF) model, a traditional dynamical downscaling approach; U-Net, a deterministic data-driven model; and Conditional Generative Adversarial Networks (CGAN), a probabilistic data-driven model. This comparison allows us to assess DWRF's capabilities in generating accurate, high-resolution forecasts that can potentially enhance renewable energy management and production.

\begin{figure}[h]
\centering
\includegraphics[width=0.9\textwidth]{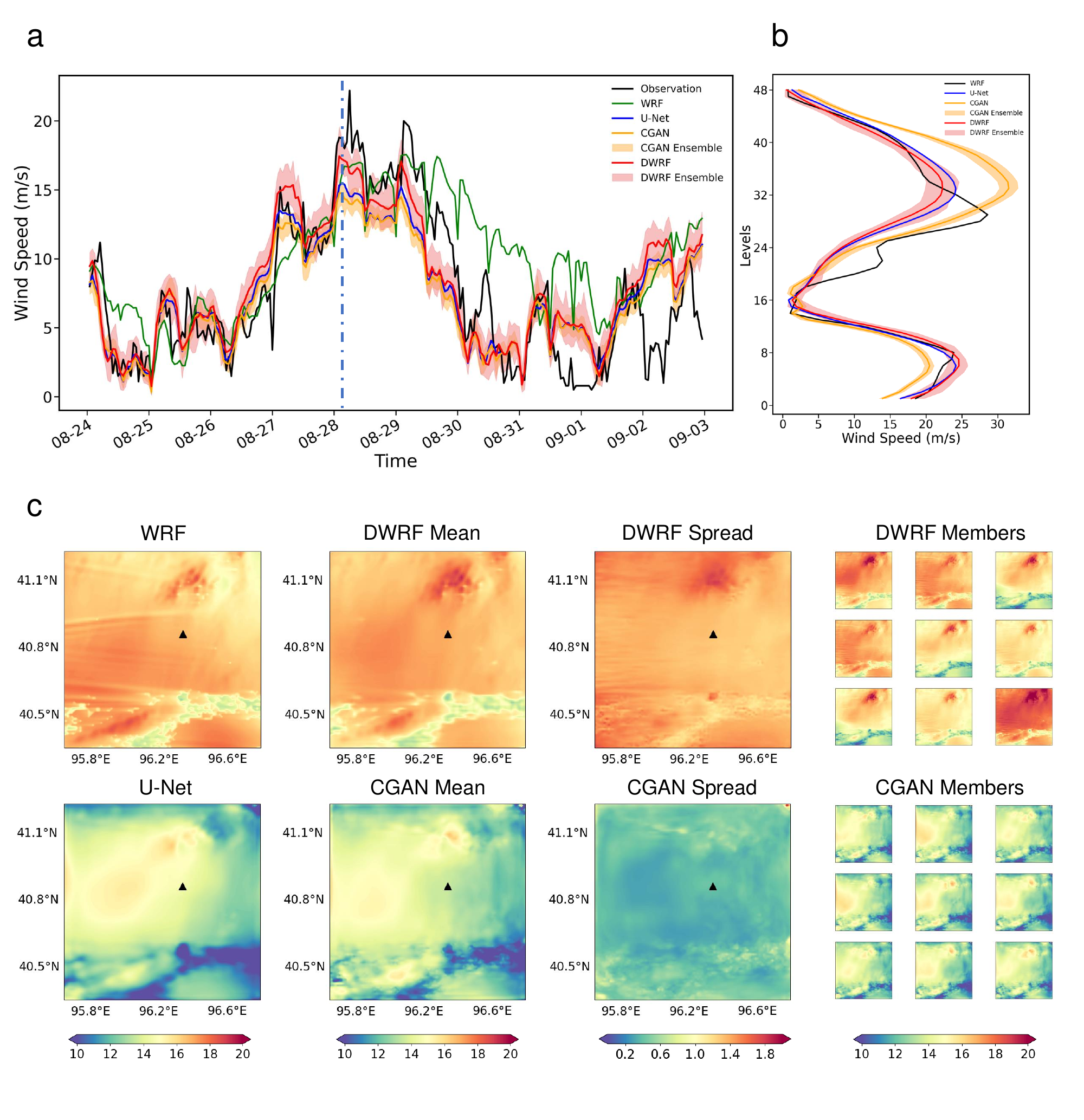}
\caption{Comparative analysis of wind speed predictions by a extreme case study. (a) Time series of 10-m wind speed forecasts over a 10-day period at the specific observation location (marked with a black triangle in panel(c)). The black line represents the "Truth" from station observations. Colored lines show predictions from various models: WRF (green), U-Net (blue), CGAN ensemble mean (yellow), and DWRF ensemble mean (red). The shaded areas indicate the ensemble spread, bounded by the minimum and maximum of the ensemble forecasts for CGAN (light pink) and DWRF (light red), respectively. (b) Vertical profiles of wind speed predictions, with the ERA5-driven WRF as reference. (c) Spatial distributions of 10-m wind speed predictions at the time marked by the blue dashed line in panel (a). The black triangle represents the observation station's location. The ensembble mean is calculated by the average of 50 members for CGAN and DWRF. The ensemble spread is calculated as the standard deviation of ensemble forecasts. 9 members are randomly plotted to show the ensemble forecasting performance.}\label{fig2}
\end{figure}

Fig. \ref{fig2} presents comparisons of 10-day wind speed predictions from different models, initialized at August 24, 2023. We first evaluate the predictions against weather station observations, as shown in Fig. \ref{fig2}a. The observation station's location is marked with a black triangle in Fig. \ref{fig2}c. The WRF model (green line) shows inconsistent predictions compared to the observed wind speed values in the first several hours. The WRF's predictions exhibit significant fluctuations and deviate from the observed pattern. This inconsistency likely stems from the spin-up problem inherent to dynamical downscaling models, caused by discrepancies between large-scale forecasts and fine-scale dynamical processes. In contrast, deep learning models, which learn the process directly from data, show better agreement with observations in the initial hours. The DWRF ensemble mean, in particular, closely follows the observed wind speed values, capturing temporal variations and magnitudes more effectively than the WRF model.

Extremely strong wind speeds were observed from August 28 to 29, a critical period for wind power production. During this time, the U-Net (blue line) and CGAN ensemble mean (orange line) models significantly underestimated wind speeds, failing to capture the higher velocities recorded at weather stations. In contrast, the DWRF ensemble mean (red line) maintained closer agreement with observed wind speed values throughout this period. It successfully captured the higher wind speeds and remained consistent with the observed pattern. This highlights the superior performance of DWRF in providing accurate wind speed predictions even in the later stages of the forecast, where other models such as U-Net and CGAN struggle to maintain accuracy.

The DWRF ensemble spread (light red shaded area) is generally wider than that of the CGAN model (light orange shaded area), indicating a more comprehensive representation of forecast uncertainty. The wider spread of the DWRF ensemble suggests it better captures the range of possible wind speed outcomes, providing a more informative measure of prediction uncertainty. As shown in Fig. \ref{fig2}a, DWRF ensemble members have better coverage of the observations, further demonstrating the model's ability to capture the full range of potential outcomes.

Vertical profiles of wind speed predictions are shown in Fig. \ref{fig2}b. Almost all models capture the general trend of increasing wind speed with altitude, including a prominent peak around levels 32-34 likely indicating a jet stream, and a secondary peak at lower levels suggestive of a low-level jet. At lower levels (0-16), CGAN tends to underestimate the wind speed. DWRF and U-Net closely follow the trend of wind predictions, with DWRF's ensemble spread providing a reliable range of predictions that more consistently encompasses the WRF profile. This highlights the DWRF model's ability to make accurate and reliable predictions at both surface and upper levels.

Fig. \ref{fig2}c shows the spatial distributions of 10-m wind speed predictions at a specific time (the blue dash line in Fig. \ref{fig2}a). Both WRF and DWRF ensemble mean reveal clearly sharp predictions, showing the effects of local terrains with fine-scale features. U-Net exhibits a more pronounced underestimation of the wind speed at this specific time step, indicating a weaker performance. Additionally, since U-Net is trained with the objective of mean squared errors (MSE), it leads to poorly structured and blurry predictions. CGAN is able to produce relatively sharp predictions compared with U-Net, since it is a probabilistic model. However, the CGAN ensemble mean also underestimates the wind speed prediction, similar to the U-Net model, which indicates the limitation of generative adversarial networks on high-resolution weather forecasts. In contrast, the DWRF ensemble mean can better capture the high-resolution spatial patterns and generate strong wind speed predictions that has the best agreement with observations. Moreover, the DWRF ensemble spread clearly reveals the forecast uncertainty, particularly in the area where strong wind speeds appear.

\begin{figure}[h]
\centering
\includegraphics[width=0.9\textwidth]{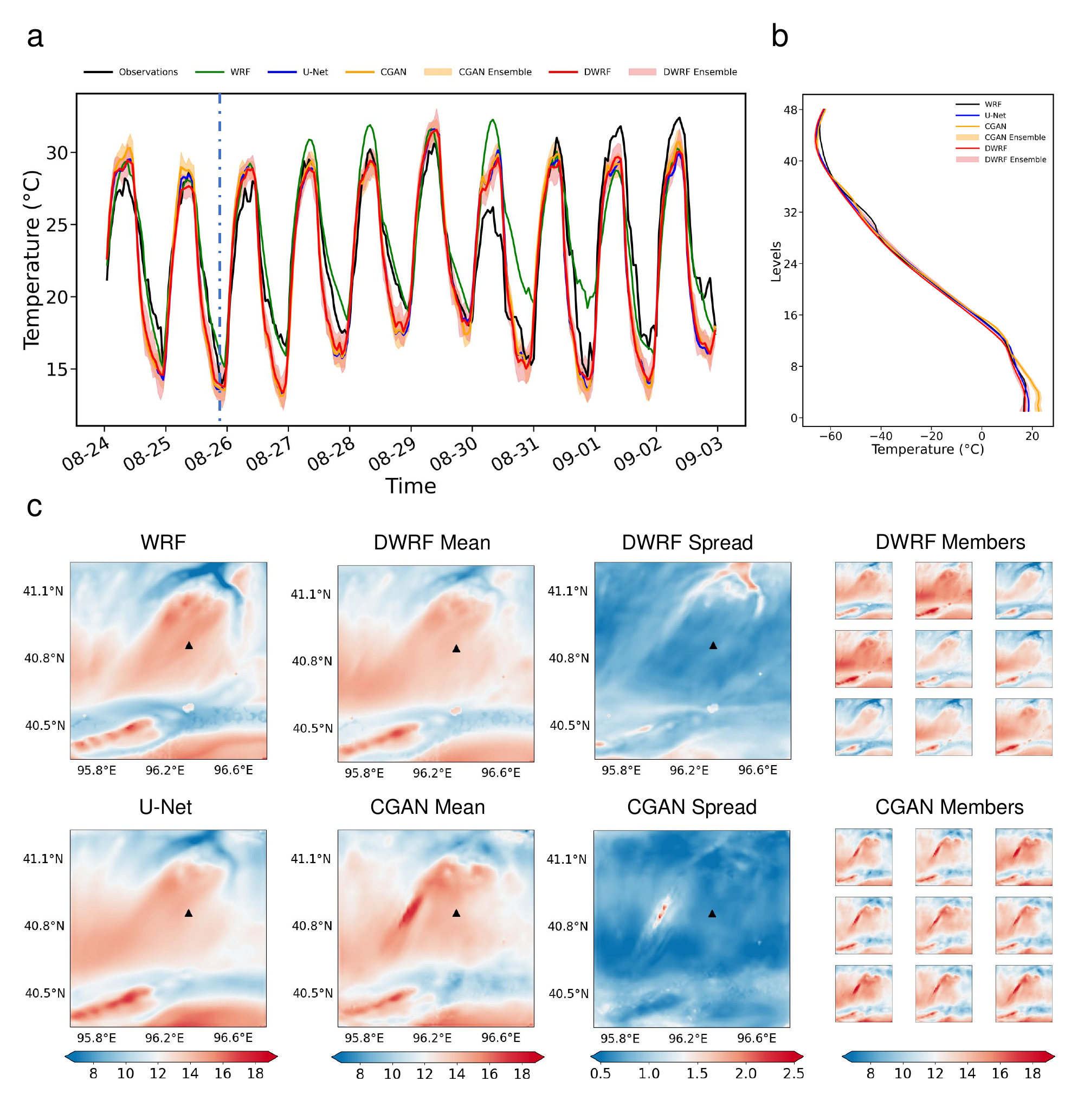}
\caption{Comparative analysis of temperature predictions. (a) Time series of 2-m temperature predictions over a 10-day period against observations. (b) Vertical profiles of temperature predictions. (c) Spatial distributions of 2-m temperature predictions from different models.}\label{fig3}
\end{figure}

Fig. \ref{fig3} presents comparisons of temperature predictions. The time series comparison of 2-m temperature predictions (Fig. \ref{fig3}a) reveals DWRF's robust performance against other models over the 10-day period. DWRF consistently demonstrates superior accuracy compared to the traditional WRF model, particularly in capturing the diurnal temperature cycle. It shows remarkable skill in predicting both daily maxima and minima, while WRF often overestimates daytime highs and underestimates nighttime lows.

The vertical profiles of temperature predictions (Fig. \ref{fig3}b) show DWRF's excellent agreement with WRF from the surface up to the highest levels. For spatial distributions (Fig. \ref{fig3}c), DWRF's mean prediction closely emulates the WRF output, accurately capturing both large-scale patterns and fine-scale features across the domain.

In summary, this case study demonstrates DWRF's superior performance in predicting both wind speed and temperature across various spatial and temporal scales. The model's ability to generate accurate mean predictions with meaningful uncertainty quantification positions DWRF as a robust tool for high-resolution meteorological forecasting, particularly valuable for wind power production services and other weather-sensitive applications.

\subsection{Forecast skills evaluation}

To comprehensively assess the performance of DWRF, we conducted an extensive evaluation over the entire year of 2023. This evaluation consisted of daily 15-day high-resolution weather forecasts, initiated at 0000 UTC each day. Fig. \ref{fig3} presents a detailed comparison of high-resolution weather forecasting performance among three models: DWRF, U-Net, and CGAN. We employed two key metrics for this evaluation: Root Mean Square Error (RMSE), which quantifies the average magnitude of forecast errors, and Continuous Ranked Probability Score (CRPS), which assesses the quality of probabilistic forecasts. 

The WRF simulation, driven by hourly ERA5 reanalysis data, served as our reference standard. Our analysis encompassed a wide range of meteorological variables, including both upper-level parameters (temperature and humidity at 500 and 700 hPa) and surface variables (10-meter wind components U10m and V10m, sea level pressure SLP, and 2-meter temperature T2m). This comprehensive approach allows us to evaluate the models' performance across different atmospheric levels and for various critical weather parameters, providing a holistic view of their forecasting capabilities.

\begin{figure}[h]
\centering
\includegraphics[width=1.0\textwidth]{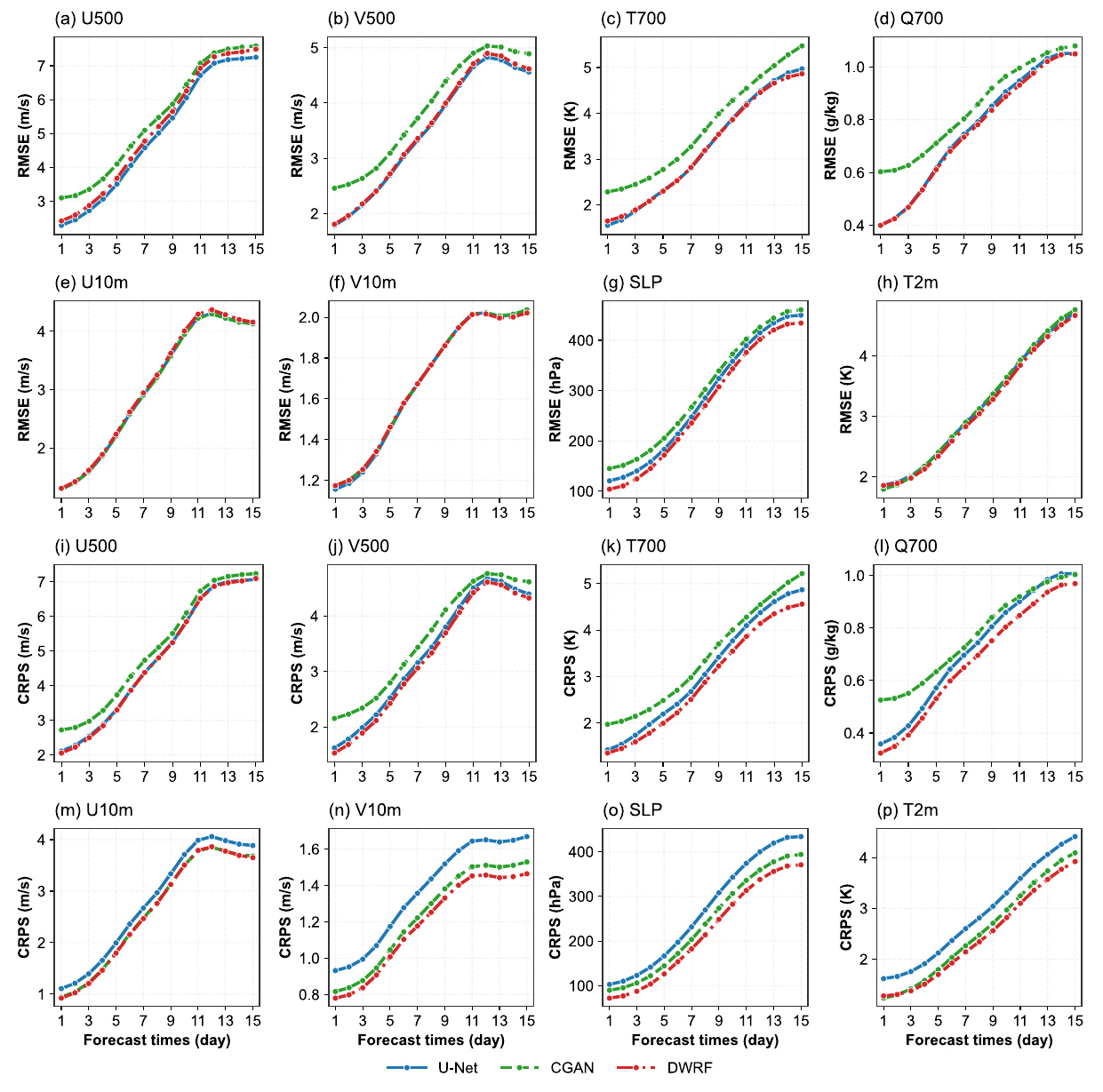}
\caption{Evaluation of high-resolution weather forecasting performance by root mean square error (RMSE) and continuous ranked probability score (CRPS) for upper-level variables (wind components at 500 hPa, temperature and humidity at 700 hPa) and surface variables (10-meter wind components, sea level pressure, and 2-meter temperature)}\label{fig4}
\end{figure}

For RMSE (top two panels), all models exhibited a consistent pattern of increasing errors as the forecast lead time extended, aligning with the expected degradation in forecast accuracy over time. For upper-level variables, we observed minimal differences between the models, suggesting comparable deterministic forecast performance at higher altitudes. Notably, the U-Net model demonstrated competitive performance, likely due to its optimization for minimizing mean squared error. However, when examining surface variables, DWRF showed markedly lower RMSE values, particularly for sea level pressure (SLP) and 2-meter temperature (T2m). This advantage became more pronounced at longer lead times, indicating DWRF's superior accuracy in predicting these critical near-surface parameters.

CRPS assesses the probabilistic forecasting performance by considering the entire probability distribution of the forecasts. As with RMSE, CRPS values increased with forecast lead time across all models, reflecting the growing uncertainty inherent in longer-range predictions. However, CRPS revealed more pronounced differences between models compared to RMSE, underscoring the importance of probabilistic forecast evaluation. Both CGAN and DWRF generally outperformed U-Net, especially for surface variables, which can be attributed to the superior uncertainty quantification capabilities of these generative models. Interestingly, CGAN seemed to struggle with upper-level variables, sometimes performing worse than U-Net, possibly due to training instabilities when dealing with high-dimensional data. In contrast, DWRF consistently achieved the lowest CRPS across all variables, with particularly impressive results for SLP and T2m. This performance highlights DWRF's proficiency in providing reliable and informative probabilistic forecasts.

In conclusion, DWRF demonstrates superior performance across both deterministic and probabilistic metrics for upper-level and surface variables. Its consistent outperformance, especially in CRPS, underscores its robust uncertainty quantification capabilities. This comprehensive skill set positions DWRF as a significant advancement in weather prediction, offering potential benefits for various weather-dependent sectors such as renewable energy industries.

\subsection{Downstream applications in wind energy planning and operation}

The superior performance of DWRF in high-resolution weather forecasting, as demonstrated in previous sections, has significant implications for renewable energy applications, particularly in the wind energy sector. Wind power has become a crucial component of the global renewable energy mix, playing a significant role in the transition towards a sustainable, low-carbon future. However, Wind power generation is heavily dependent on accurate weather predictions, as wind resources are inherently variable and intermittent, influenced by complex atmospheric dynamics and local weather conditions \cite{wang2021review}. In this subsection, we evaluate how DWRF's enhanced forecasting capabilities translate into practical benefits for wind energy planning and operations.

\begin{figure}[h]
\centering
\includegraphics[width=0.95\textwidth]{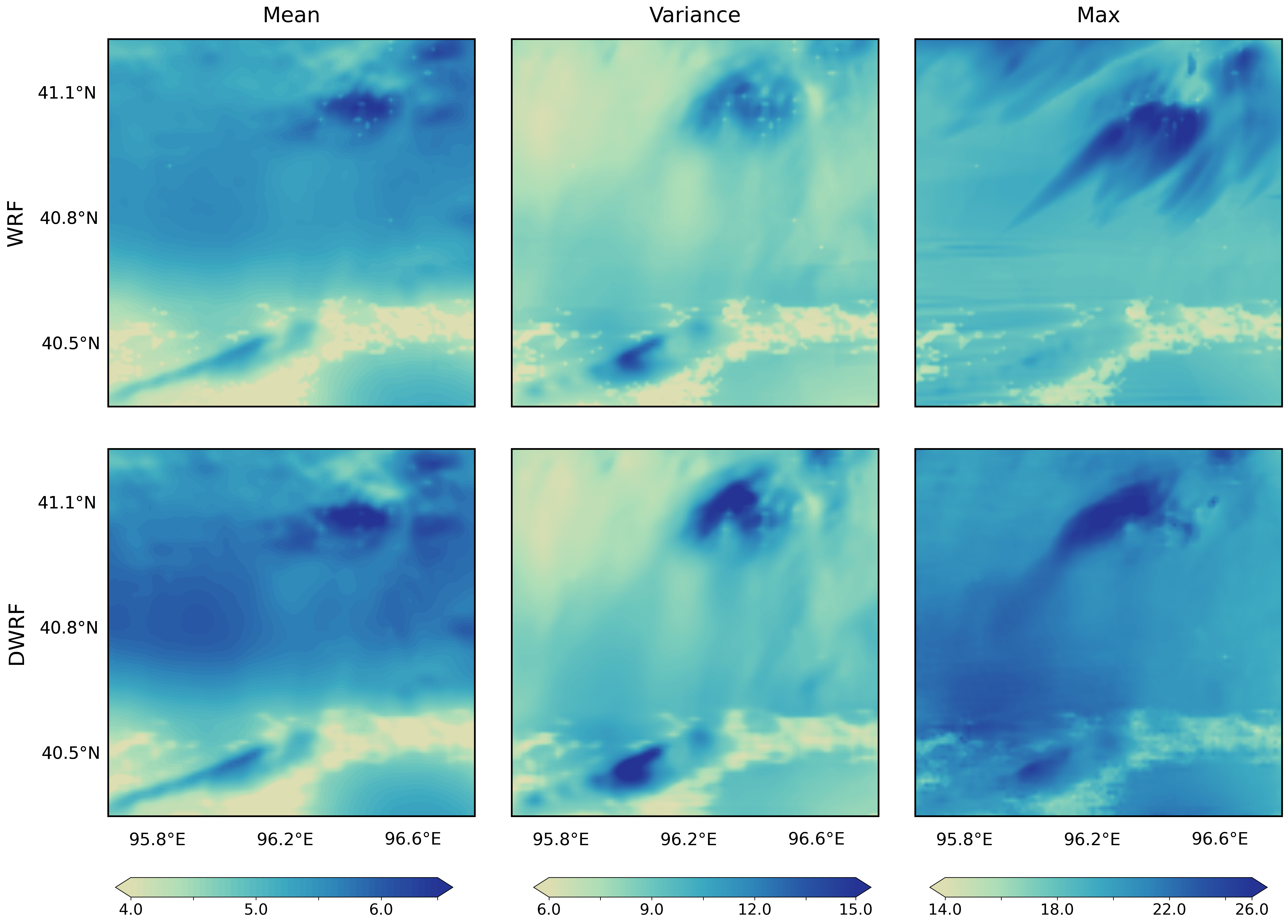}
\caption{Comparison of climatological statistics for 10m wind speed predictions between WRF (top row) and DWRF (bottom row). The mean (left column), variance (middle column), and maximum (right column) statistics are calculated across 3000 samples for each grid point in the domain. For WRF, samples were randomly selected from model output, while for DWRF, samples were generated using the unconditional diffusion-based model.
}\label{fig5}
\end{figure}

To assess DWRF's potential in long-term wind energy planning, we first compare its ability to generate climatological statistics with the established WRF model. Fig.\ref{fig5} shows the comparison of climatological statistics for 10m wind speed predictions between WRF and DWRF. We randomly selected 3000 samples for both models and calculated the mean, variance, and maximum 10m wind speed across all samples. The results demonstrate strong agreement between the two models, with both showing similar spatial patterns in mean wind speed, variance, and maximum wind speed distributions. This consistency validates DWRF's ability to accurately capture the climatological characteristics of the study area, matching the established performance of WRF.

DWRF's key advantage lies in its computational efficiency for generating climatological statistics. Unlike the computationally intensive WRF, DWRF can rapidly produce numerous samples, enabling comprehensive analysis that would be prohibitively expensive with WRF alone. This efficiency is crucial for wind energy planners who need to quickly assess multiple potential sites or scenarios. The mean wind speed distribution helps identify optimal locations for wind plant placement, targeting areas with consistently high wind speeds. The variance maps inform about wind variability, with preference given to areas of lower variance to ensure more stable power output. Maximum wind speed distributions highlight regions prone to extreme winds, which should be avoided to prevent potential damage to wind turbines.

In summary, DWRF's ability to efficiently generate accurate climatological statistics makes it a valuable tool for the wind energy sector. It offers the potential for more extensive and detailed analyses than WRF, contributing significantly to optimizing wind energy production, enhancing infrastructure resilience, and improving operational efficiency in renewable energy.

Besides wind plant siting and planning, high-resolution forecasts from DWRF help improve wind energy operations. We assess its performance in wind power prediction. We ran wind power forecasting models driven by different forecasting products from U-Net, CGAN, and DWRF. Wind power forecasting models consists of ensemble methods, including CNN, CNN-LSTM, ED-LSTM, ED-GRU, Light Gradient Boosting Machine (LightGBM), and eXtreme Gradient Boosting (XGBoost). The study area included 200 wind plants as reference, with all models using wind speed, wind direction, temperature, and humidity at 90 meters height as input to predict wind power output.

\begin{figure}[h]
\centering
\includegraphics[width=0.95\textwidth]{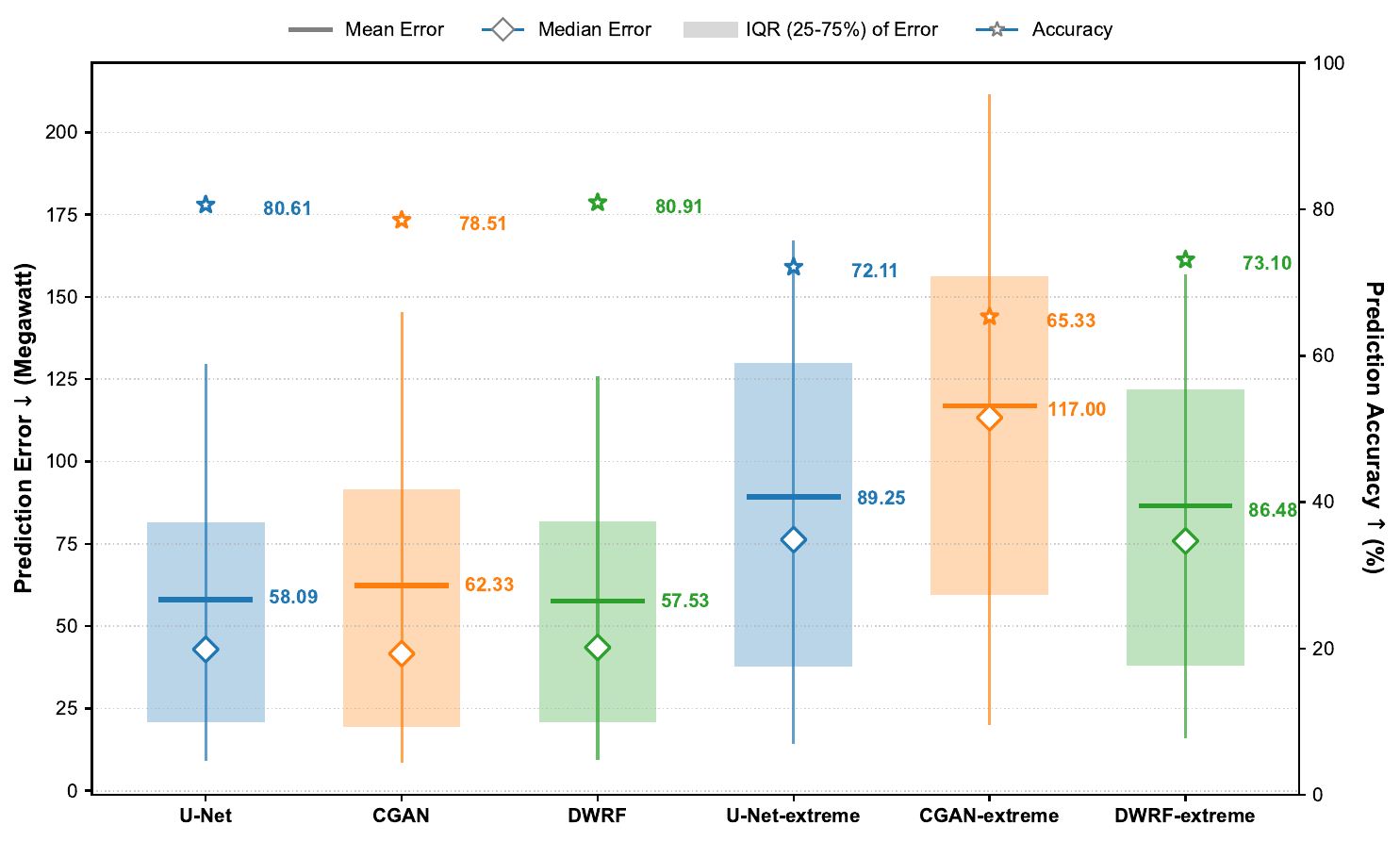}
\caption{Comparison of wind power prediction error (Megawatt) and prediction accuracy ($\%$) driven by different forecasting products.}\label{fig7}
\end{figure}

Fig. \ref{fig7} presents the performance evaluation encompassing two scenarios: average performance over a six-month period in 2023 and performance for extreme cases (defined as wind power exceeding 173 MW, which occurs in 25$\%$ of the historical data). The results demonstrate that models driven by DWRF forecasts consistently achieve lower prediction errors, and higher prediction accuracy ($C_R$, definition seen Subsection \ref{sec_eva}) values compared to U-Net and CGAN. DWRF's performance is particularly notable in extreme cases, indicating robust predictive capability under challenging circumstances.

Focusing on prediction accuracy, which indicates wind power prediction accuracy, DWRF improves upon U-Net by 0.3$\%$ and CGAN by 2.4$\%$ on average. In extreme cases, these improvements are more pronounced, with DWRF enhancing prediction accuracy by 1.177$\%$ and 8.096$\%$ compared to U-Net and CGAN, respectively. While these improvements may seem modest, they translate into substantial economic benefits. For instance, assuming an electricity price of $\$$100 per MWh, a 1.177$\%$ improvement in prediction accuracy could result in a financial benefit of $\$$2,988.403 per day for the target wind farm. This enhancement enables better optimization of operations, reduction of balancing costs, and potential increase in revenue.

These findings underscore the importance of fine-grained spatial and temporal information captured by DWRF forecasts in accurately representing local weather conditions and their influence on wind power generation. The ability to more precisely predict wind power output, particularly during extreme events, has significant implications for grid stability, economic efficiency, and the overall integration of wind energy into power systems.

In conclusion, DWRF represents a competitive advance in weather modeling for wind energy applications. Its ability to efficiently generate accurate climatological statistics offers a robust foundation for wind plant siting decisions in the planning phase. For day-to-day operations, DWRF's superior short-term forecasting capabilities translate into more accurate wind power predictions, particularly during extreme events. As the wind energy sector continues to grow and face new challenges, models like DWRF will play a crucial role in optimizing resource utilization, improving grid stability, and ultimately contributing to a more sustainable energy future.

\section{Discussion}\label{sec3}

The DWRF model introduced in this study represents a significant advancement in high-resolution ensemble weather forecasting, demonstrating superior performance compared to traditional and other machine learning approaches. Our results indicate that DWRF not only matches the accuracy of the WRF model but also substantially outperforms it in terms of computational efficiency. By leveraging the power of deep learning and GPU acceleration, our model yields a 150-member 10-day forecast, with 1-km and 15-min resolution, within 1hour on a NVIDIA GeForce RTX 4090 GPU, while the WRF model requires approximately 30 hours on 40 CPUs to produce the same forecast. This significant reduction in computational time and resources highlights the efficiency gains achieved by the DWRF model, making it a promising solution for operational high-resolution weather forecasting.

When compared to other machine learning models, DWRF shows distinct advantages. Unlike CNN-based models such as U-Net, which tend to produce blurred outputs at high resolutions, DWRF generates crisp, well-defined patterns that capture intricate spatial details. This capability is crucial for accurately representing fine-scale atmospheric phenomena, which are particularly important in applications like wind energy forecasting. Moreover, DWRF's performance in uncertainty quantification surpasses that of Conditional Generative Adversarial Networks (CGANs), offering more reliable probabilistic forecasts. This enhanced ability to represent forecast uncertainty is invaluable for risk assessment and decision-making in weather-sensitive industries.

The DWRF model's adaptability is another key strength that sets it apart from other approaches. Trained on a high-resolution weather simulation dataset in a self-supervised manner, the DWRF model can be easily coupled with various regression models at the inference stage. This flexibility allows the DWRF model to adapt to different large-scale data sources, making it highly versatile for operational use where input data quality and availability may vary. However, it's important to note that this adaptability may come with challenges in ensuring consistent performance across different data sources and geographical regions, which warrants further investigation.

\section{Conclusion}\label{sec4}
This study introduces DWRF (Diffusion-based Weather Research and Forecasting), as an innovative solution to the growing demand for accurate, high-resolution weather predictions, particularly in the renewable energy sector. By integrating a large-scale global weather forecasting model (Pangu) with a diffusion-based probabilistic downscaling approach, DWRF addresses the limitations of traditional numerical weather prediction models while providing computationally efficient, high-resolution ensemble forecasts.

The key contributions of DWRF lie in its ability to generate spatially and temporally coherent atmospheric fields at 1-km and 15-minute resolutions, its superior performance in capturing fine-scale patterns, and its robust uncertainty quantification through ensemble forecasting, with great computational efficiency. These capabilities position DWRF as a powerful tool for advancing weather-dependent decision-making across multiple industries.

In the context of wind energy, DWRF demonstrates significant potential to enhance both long-term planning and day-to-day operations. Its efficient generation of accurate climatological statistics provides a solid foundation for optimal wind plant siting. Furthermore, DWRF's improved wind power prediction accuracy, especially during extreme events, translates to tangible economic benefits and more efficient grid integration.

As the demand for precise, high-resolution weather forecasts continues to grow across various sectors, DWRF stands poised to make significant contributions. Its development represents a step forward in our ability to understand and predict complex atmospheric processes, ultimately supporting more informed decision-making and contributing to the sustainable management of weather-sensitive resources and industries.

While DWRF shows great promise, there are several potential improvements to consider in the future. First, investigating DWRF's transferability to diverse geographical domains and climate regimes will be crucial for its widespread adoption. Second, enhancing the model's scalability to cover larger areas without compromising resolution or efficiency could further expand its applicability. Lastly, integrating DWRF with sector-specific models and decision support systems could unlock its full potential in areas such as renewable energy, agriculture, and urban planning.

\section{Methods}\label{sec5}
\subsection{Dataset}
We selected an operational wind plant as the target region and generated a high-resolution meteorological simulation dataset spanning a period of 2 years. The dataset comprises data at a spatial resolution of 1 km and a temporal resolution of 15 minutes, created using the Advanced Research Version 4.5 of the Weather Research and Forecasting (WRF-ARW V4.5) model \cite{powers2017weather}. The study area is set within the geographical bounds of 40.2-41.2°N and 95.6-96.8°E. To accommodate this region, the WRF model was configured with a two-nested domain using the Lambert conformal conical projection method. The outer domain has a resolution of 3 km with a grid of $170 \times 173$ points, while the inner domain operates at a higher resolution of 1 km with a grid of $106 \times 109$ points. Vertically, both domains consist of 51 sigma layers, with the model top set at 50 hPa. The initial and boundary conditions for the WRF simulation were derived from the fifth generation of the European Centre for Medium-Range Weather Forecasts (ECMWF) atmospheric reanalysis data (ERA5) \cite{hersbach2020era5}. To improve the accuracy of the simulation, we employed the Analysis Nudging technique, which introduces forcing terms to the governing equations, allowing the model output to be adjusted towards the reanalysis data. The WRF simulation was run at 0000 UTC every 5 days, covering the period from September 2021 to September 2023, with an output frequency of 15 minutes.

In addition to driving the WRF simulation, the ERA5 reanalysis dataset was utilized as the low-resolution input for training the downscaling model in this study. The ERA5 dataset has a spatial resolution of approximately 25 km and a temporal resolution of 1 hour. To encompass a larger area than the target domain, we cropped the ERA5 dataset into a grid of $41 \times 61$ pixels. As the low-resolution input, we selected five variables: geopotential height, specific humidity, temperature, and the eastward and northward components of the wind vector. These variables were considered at 13 pressure levels: 50 hPa, 100 hPa, 150 hPa, 200 hPa, 250 hPa, 300 hPa, 400 hPa, 500 hPa, 600 hPa, 700 hPa, 850 hPa, 925 hPa, and 1,000 hPa.

\subsection{Diffusion probabilistic model}\label{method-diffusion}
Diffusion model is a deep generative model that learns the probability distribution of the target variable. Diffusion model defines a stochastic process that translates the target distribution into standard Gaussian distribution, and learns a series of variational distributions to reverse this process, hence achieving generative modeling. This methodology offers extreme flexibility to learn high-dimensional distributions, and guarantees effective control of conditioning information over the generated distribution. Also, since diffusion models explicit likelihood based models, we do not suffer from mode collapse in distribution estimation, ensuring reliable probabilistic forecast. 

In diffusion probabilistic model, we define the following stochastic process that turns a target distribution p(${x}_{0}$) into standard Gaussian. Here ${x}_{0}$ is random variable of a target meteorological field, for instance, 10m wind velocity over the target region. 

\begin{equation}
q\left(\mathbf{x}_{t} \mid \mathbf{x}_{0}\right)=\mathcal{N}\left(\alpha_{t} \mathbf{x}_{0}, \sigma_{t}^{2} \mathbf{I}\right)
\end{equation}

Here ${x}_{t}$ is latent variable indexed by $t \in[0,T]$, $\alpha_{t}/\sigma_{t}$ is monotonically decreasing/increasing function of t, bounded by [0, 1]. This stochastic process therefore bridges p(${x}_{0}$) and $\mathcal{N}\left(0,\mathbf{I}\right)$. We reverse this process to transform $\mathcal{N}\left(0,\mathbf{I}\right)$ into p(x), using a chain of variational distributions.

\begin{equation}
p_{\theta}\left(\mathbf{x}_{t-1} \mid \mathbf{x}_{t}\right)=\mathcal{N}\left(\mu_{\theta}\left(\mathbf{x}_{t}\right), {\Sigma}_{\theta}\left(\mathbf{x}_{t}\right)\right), \quad t \in[1, T]
\end{equation}

Here $t$ is arbitrary discretizations of time; $\left\{\mu_{\theta}, {\Sigma}_{\theta}\right\}$ are learnable mean vector and covariance matrix, trained by maximizing the overall data likelihood. To achieve tractable optimization, we consider the following factorization of data likelihood:

\begin{equation}
\begin{aligned}
    \log p_{\theta}(\mathbf{x}) = & \ \mathbb{E}_{q\left(\mathbf{x}_{1} \mid \mathbf{x}_{0}\right)} \log p_{\theta}\left(\mathbf{x}_{0} \mid \mathbf{x}_{1}\right) \\
    & - \mathbb{E}_{q\left(\mathbf{x}_{T-1} \mid \mathbf{x}_{0}\right)} D_{\mathrm{KL}}\left(q\left(\mathbf{x}_{T} \mid \mathbf{x}_{T-1}\right) \| p\left(\mathbf{x}_{T}\right)\right) \\
    & - \sum_{i=1}^{T-1} \mathbb{E}_{q\left(\mathbf{x}_{t} \mid \mathbf{x}_{0}\right)} D_{\mathrm{KL}}\left(q\left(\mathbf{x}_{t} \mid \mathbf{x}_{t-1}\right) \| p_{\theta}\left(\mathbf{x}_{t} \mid \mathbf{x}_{t+1}\right)\right)
\end{aligned}
\end{equation}

To maximize the equation above is approximately equivalent to minimizing the Fisher divergence between the data and model distributions:

\begin{equation}
\theta^*=\arg \max _{\theta} \log p_\theta(\mathbf{x}) \approx \arg \min _{\theta} \sum_{i=1}^T \mathbb{E}_{p\left(\mathbf{x_i} \mid \mathbf{x_0}\right)}\left\|\nabla \log p\left(\mathbf{x_i} \mid \mathbf{x_0}\right)-\epsilon_{\mathrm{NN}_\theta}\left(\mathbf{x}_{i}\right)\right\|_2
\end{equation}

Here $\epsilon_{\mathrm{NN}_\theta}$ is a neural network parameterization of $\log p\left(\mathbf{x_t} \mid \mathbf{x_0}\right)$, known as the score function. Given the trained score estimates, we can derive $p_{\theta}\left(\mathbf{x_{t-1}} \mid \mathbf{x_t}\right) = \mathcal{N}\left(\mu_{\theta}\left(\mathbf{x_t}\right), {\Sigma}_{\theta}\left(\mathbf{x_t}\right)\right)$ and sample it, starting with $p_{\theta}\left(\mathbf{x_T}\right)= \mathcal{N}\left(0, 1\right)$, ending with $p\left(\mathbf{x_0} \right)$.

\subsection{SDEdit-based conditional sampling}
The diffusion model demonstrates remarkable performance in learning the probabilistic prior distribution of the target variable fields through a self-supervised learning approach. The primary objective of this study is to generate high-resolution atmospheric fields ($x$) based on the conditions provided by low-resolution data ($y$). In essence, we aim to approximate the conditional distribution $p\left(\mathbf{x} \mid \mathbf{y}\right)$.

Directly including the condition $y$ necessitates retraining the model, which is computationally expensive given the time and resources required to train a diffusion model \cite{song2021solving}. Moreover, this approach may be suboptimal due to the presence of uncertainty associated with different sets of information. Instead, we introduce the condition information at the inference stage while preserving the pretrained unconditional diffusion model as the foundation model. To achieve this, we preprocess the training dataset into pairs, $\left({x_i, y_i}\right)$, where y and x represent the low-resolution and the corresponding high-resolution data, respectively. We then train a CNN-based regression model ($f_{\theta^\ast}$) to downscale the input large-scale fields in a supervised manner, as the preliminary prediction results (Fig. \ref{fig1}a), $f_{\theta^\ast}\left({y}\right)$. As discussed in the Introduction, training a supervised task with the objective function of Mean Squared Error (MSE) tends to learn the mean vector, assuming data follow Gaussian distributions. 

 

Recall that with the unconditional diffusion model, we have learned a climatological prior which estimates the probabilistic distribution of high-resolution atmospheric states. We also have a deterministic regression model that produce preliminary high-resolution prediction from low-resolution atmospheric fields. We use Stochastic Differential Equation Editing (SDEdit)\cite{meng2021sdedit} to produce accurate and reliable high-resolution weather forecasts. Instead of generating state estimates from pure noise, we start the diffusion process from the preliminary prediction with carefully calibrated noise.

The SDEdit framework leverages the learned climatological prior with the preliminary prediction through the following two-step process.
 \begin{enumerate}[i)]
\item \textbf{Forward diffusion:} add Gaussian noise to the preliminary prediction of the regression model, up to a certain noise level ($t_0$), as is shown in Fig. \ref{fig1}a.
\item \textbf{Reverse diffusion}: run the reverse sampling process from the noisy preliminary prediction ($t_0$), progressively removing noise and generate the final prediction output. 
\end{enumerate}

\subsection{Optimal noise level in SDEdit}

To determine the optimal noise level $t_0$, we analyze the expected error between the true state $\mathbf{X}_t$ and the SDEdit estimate $\text{SDEdit}(f_{\theta^\ast}\left({y}\right), t_0)$. The squared $L_2$ error can be bounded as:

\begin{equation}
\begin{split}
    \|\mathbf{X}_{t+\Delta t} - \text{SDEdit}(f_{\theta^\ast}\left({y}\right), t_0)\|_2^2 &\leq \, \|\mathbf{X}_{t+\Delta t} - \text{SDEdit}(\mathbf{X}_{t+\Delta t}, \tau^*)\|_2^2 \\
    &+ \|\text{SDEdit}(\mathbf{X}_{t+\Delta t}, \tau^*) - \text{SDEdit}(f_{\theta^\ast}\left({y}\right), t_0)||_2^2
\end{split}
\end{equation}

This inequality decomposes the prediction error into two components: inherent information loss from applying noise to the true state (left), and how differences between the true state and the preliminary prediction through the diffusion process (right), with impacts balanced by the critical parameter $t_0$.
 
\begin{enumerate}[i)]
\item For small values of $t_0$ and insufficient noise, the distribution remains tightly constrained around $f_{\theta^\ast}\left({y}\right)$, preserving the systematical biases in $f_{\theta^\ast}$;
\item For large values of $t_0$ and excessive noise, the distribution approaches the unconstrained climatological prior, losing valuable condition information.
\end{enumerate}
 
 Thus, the parameter that how much noise we add into the preliminary prediction serves as a hyperparameter that strikes a balance between the influence of the condition information and the prior distribution learned by the unconditional diffusion model. We need to find a proper initialization from which we can solve the reverse SDE to obtain accurate and reliable weather forecasts. 
 
\begin{figure}[h]
\centering
\includegraphics[width=0.95\textwidth]{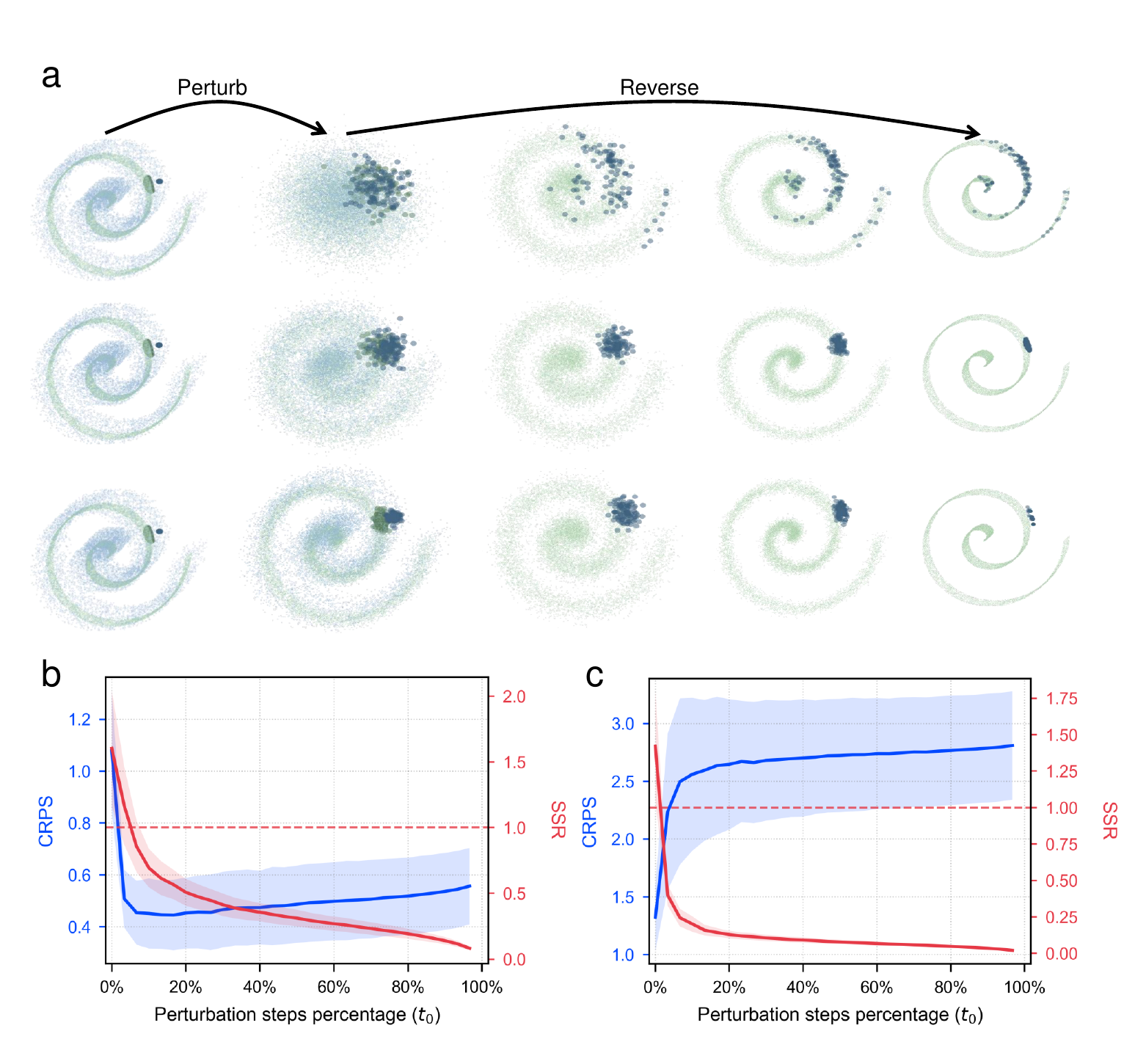}
\caption{Influence of the perturbation steps on the DWRF performance. (a) Schematic representation of the DWRF process. The process begins with an initial state (leftmost column), undergoes perturbation (second column from left), and then goes through multiple reverse steps (three rightmost columns) into the final output. The three rows shows different scenarios when too much noises (top), proper amount of noise (middle) and too little noise are added to the preliminary prediction. (b) and (c) show performance metrics of the DWRF model as a function of the perturbation steps ($t_0$). (b) illustrates the situation when the preliminary prediction is relatively consistent with the ground truth and (c) shows the one when the preliminary prediction fall far behind of the ground truth. The blue line and shaded area represent the Continuous Ranked Probability Score (CRPS) and its uncertainty range, respectively. The red line and shaded area show the Spread-Skill Ratio (SSR) and its uncertainty range. The dashed red line indicates the optimal SSR value of 1.0.
}\label{fig6}
\end{figure}

 Fig. \ref{fig6}a illustrates the influence of three different perturbation steps ($t_0$) on the DWRF performance. Fig. \ref{fig6}b and Fig. \ref{fig6}c show the CRPS and SSR as a function of perturbation steps. In Fig. \ref{fig6}b, the preliminary prediction is relatively close to ground truth. Initially, at $t_0$ = 0$\%$, we observe high CRPS, SSR above 1, indicating poor forecast skill with an overdispersed ensemble. This corresponds to the first row in Fig. \ref{fig6}a when we add too much noise into the preliminary prediction that the reverse process nearly begins from a pure random Gaussian noise. In that case, DWRF only use its climatology prior information without conditioned on the preliminary prediction, leading to low prediction accuracy and overestimated ensemble spread (diverse blue dots in the rightmost column on the first row). As $t_0$ increases to around 10$\%$, we see a rapid improvement: CRPS drops sharply, indicating better forecast accuracy; SSR approaches 1, suggesting an appropriate spread-skill relationship. We find that adding proper amount of noise will lead to better prediction (middle row in Fig. \ref{fig6}a). This range of $t_0$ appears to offer an optimal balance between accuracy and spread. However, as $t_0$ continues to increase beyond this point, we observe a gradual degradation in ensemble properties without significant improvements in forecast skill. CRPS remains relatively stable with a slight upward trend, while SSR continues to decrease well below 1, indicating growing underdispersion. These trends suggest that larger perturbations lead to an increasingly narrow ensemble spread that fails to capture the full range of possible outcomes. In Fig. \ref{fig6}c, when the preliminary results fail to capture the main pattern of prediction, we observe a different trend of metrics. The CRPS continue to increase as $t_0$ increase, indicating that a bad preliminary prediction leads to a poor prediction, even worse than the climatology ($t_0$=0$\%$). 
 
 In all, this analysis underscores the critical importance of carefully tuning the perturbation step to achieve ensemble forecasts that optimally balance accuracy with a realistic representation of forecast uncertainty. These insights can help in fine-tuning the perturbation step to achieve a balance between forecast accuracy, calibration, and uncertainty representation.

\subsection{Network architectures and training details}
We employ the U-Net architecture \cite{ronneberger2015u} as the backbone of our unconditional diffusion model. The UNet embodies an encoder-decoder structure featuring skip connections between corresponding encoder and decoder layers. The encoder section downsamples input images via a sequence of convolutional layers and pooling operations, extracting hierarchical features. Conversely, the decoder section employs transposed convolutions to upsample encoded features back to the original resolution, while integrating skip connections from the encoder to recover spatial details. Our configuration comprises 4 layers in both encoder and decoder, with the base channel set to 128, doubling with each layer increase. Each UNet block integrates convolutional layers, group normalization, and SiLU activation functions, facilitating feature extraction and preservation. Furthermore, sinusoidal position embedding is integrated to capture temporal information within the diffusion process.

During the condition-guided reverse process, we establish a regression model to make preliminary prediction for high-resolution atmospheric fields. The preliminary predictions provide condition information for the reverse process in diffusion model. The regression model adopts a 3D CNN architecture with 3 layers in both the encoder and decoder components. The input volume undergoes successive 3D convolutional operations, followed by batch normalization and SiLU activation, to extract hierarchical features while reducing spatial dimensions via max-pooling. Subsequently, the decoder employs transposed convolutions to upsample the encoded features back to the original resolution. Noteworthy, the architecture incorporates specific functionalities for padding and cropping 3D volumes to ensure compatibility across various resolution levels.

To train our models, we employed the Adam optimizer \cite{kingma2014adam} with an initial learning rate of $1 \times 10^{-4}$. Additionally, we implemented the Cosine Annealing learning rate scheduler, which cyclically adjusts the learning rate over 20 epochs, gradually decreasing it from the initial rate to a minimum value (0.01 times the initial rate). This approach effectively prevents the model from becoming trapped in local minima by allowing it to explore a wider range of the loss landscape during training.

\subsection{Evaluation metrics}\label{sec_eva}
In this study, we employs several metrics to evaluate the performance of all models, including deterministic and probabilistic metrics. The specific definitions are given as follows, which are mean absolute error (MAE), root mean square error
(RMSE), wind power prediction accuracy (corrected RMSE, as$C_R$), and continuous ranked probability score (CRPS).

\begin{equation}
\text{MAE} = \frac{1}{n} \sum_{i=1}^{n} \left| y_i - \hat{y_i} \right|
\end{equation}

\begin{equation}
\text{RMSE} = \sqrt{\frac{1}{n} \sum_{i=1}^{n} (y_i - \hat{y}_i)^2}
\end{equation}

\begin{equation}
\text{$C_R$} = 1-\sqrt{\frac{1}{n} \sum_{i=1}^{n} (\frac{y_i - \hat{y}_i}{C_i})^2}
\end{equation}

\begin{equation}
CRPS(F, y) = \int_{-\infty}^{\infty} \left( F(z) - \mathbf{1}\left(z \geq y\right) \right)^2 dz
\end{equation}

\begin{equation}
F(z) = \frac{1}{M} \sum_{m=1}^{M}\mathbf{1}\left(x_m \leq z\right)
\end{equation}

\begin{equation}
SSR = \frac{\sqrt{\frac{1}{n} \sum_{i=1}^{n} (\hat{y}_i - \bar{y})^2}}{\sqrt{\frac{1}{n} \sum_{i=1}^{n} (y_i - \hat{y}_i)^2}}
\end{equation}

MAE and RMSE measures the deterministic performance, where $\hat{y}_i$ is the prediction, $\bar{y}$ is the ensemble mean of prediction, $y_i$ is the ground truth, and $n$ represents the number of samples. In ensemble forecasts, the prediction is calculated by the ensemble mean for MAE and RMSE. $C_R$ is used to evaluate the accuracy of wind power prediction, where $C_i$ is the $i$th startup capacity of the wind farm. CRPS shows the performance of probabilistic forecasts, where $F(z)$ is the CDF of the predictive distribution and $\mathbf{1}\left(z \geq y\right)$ represents the indicator function, which is 1 if $z \geq y$ and 0 otherwise. $M$ is the number of ensemble members.

\bibliography{main_bibliography}


\begin{thebibliography}{44}
\ifx \bisbn   \undefined \def \bisbn  #1{ISBN #1}\fi
\ifx \binits  \undefined \def \binits#1{#1}\fi
\ifx \bauthor  \undefined \def \bauthor#1{#1}\fi
\ifx \batitle  \undefined \def \batitle#1{#1}\fi
\ifx \bjtitle  \undefined \def \bjtitle#1{#1}\fi
\ifx \bvolume  \undefined \def \bvolume#1{\textbf{#1}}\fi
\ifx \byear  \undefined \def \byear#1{#1}\fi
\ifx \bissue  \undefined \def \bissue#1{#1}\fi
\ifx \bfpage  \undefined \def \bfpage#1{#1}\fi
\ifx \blpage  \undefined \def \blpage #1{#1}\fi
\ifx \burl  \undefined \def \burl#1{\textsf{#1}}\fi
\ifx \doiurl  \undefined \def \doiurl#1{\url{https://doi.org/#1}}\fi
\ifx \betal  \undefined \def \betal{\textit{et al.}}\fi
\ifx \binstitute  \undefined \def \binstitute#1{#1}\fi
\ifx \binstitutionaled  \undefined \def \binstitutionaled#1{#1}\fi
\ifx \bctitle  \undefined \def \bctitle#1{#1}\fi
\ifx \beditor  \undefined \def \beditor#1{#1}\fi
\ifx \bpublisher  \undefined \def \bpublisher#1{#1}\fi
\ifx \bbtitle  \undefined \def \bbtitle#1{#1}\fi
\ifx \bedition  \undefined \def \bedition#1{#1}\fi
\ifx \bseriesno  \undefined \def \bseriesno#1{#1}\fi
\ifx \blocation  \undefined \def \blocation#1{#1}\fi
\ifx \bsertitle  \undefined \def \bsertitle#1{#1}\fi
\ifx \bsnm \undefined \def \bsnm#1{#1}\fi
\ifx \bsuffix \undefined \def \bsuffix#1{#1}\fi
\ifx \bparticle \undefined \def \bparticle#1{#1}\fi
\ifx \barticle \undefined \def \barticle#1{#1}\fi
\bibcommenthead
\ifx \bconfdate \undefined \def \bconfdate #1{#1}\fi
\ifx \botherref \undefined \def \botherref #1{#1}\fi
\ifx \url \undefined \def \url#1{\textsf{#1}}\fi
\ifx \bchapter \undefined \def \bchapter#1{#1}\fi
\ifx \bbook \undefined \def \bbook#1{#1}\fi
\ifx \bcomment \undefined \def \bcomment#1{#1}\fi
\ifx \oauthor \undefined \def \oauthor#1{#1}\fi
\ifx \citeauthoryear \undefined \def \citeauthoryear#1{#1}\fi
\ifx \endbibitem  \undefined \def \endbibitem {}\fi
\ifx \bconflocation  \undefined \def \bconflocation#1{#1}\fi
\ifx \arxivurl  \undefined \def \arxivurl#1{\textsf{#1}}\fi
\csname PreBibitemsHook\endcsname

\bibitem[\protect\citeauthoryear{Wiatros-Motyka et~al.}{2024}]{wiatros2024global}
\begin{botherref}
\oauthor{\bsnm{Wiatros-Motyka}, \binits{M.}},
\oauthor{\bsnm{Fulghum}, \binits{N.}},
\oauthor{\bsnm{Jones}, \binits{D.}}:
Global electricity review 2024.
Ember, United Kingdom
(2024)
\end{botherref}
\endbibitem

\bibitem[\protect\citeauthoryear{Halloran et~al.}{2024}]{halloran2024data}
\begin{barticle}
\bauthor{\bsnm{Halloran}, \binits{C.}},
\bauthor{\bsnm{Lizana}, \binits{J.}},
\bauthor{\bsnm{Fele}, \binits{F.}},
\bauthor{\bsnm{McCulloch}, \binits{M.}}:
\batitle{Data-based, high spatiotemporal resolution heat pump demand for power system planning}.
\bjtitle{Applied Energy}
\bvolume{355},
\bfpage{122331}
(\byear{2024})
\end{barticle}
\endbibitem

\bibitem[\protect\citeauthoryear{Agupugo et~al.}{2024}]{agupugo2024optimization}
\begin{barticle}
\bauthor{\bsnm{Agupugo}, \binits{C.P.}},
\bauthor{\bsnm{Kehinde}, \binits{H.M.}},
\bauthor{\bsnm{Manuel}, \binits{H.N.N.}}:
\batitle{Optimization of microgrid operations using renewable energy sources}.
\bjtitle{Engineering Science \& Technology Journal}
\bvolume{5}(\bissue{7}),
\bfpage{2379}--\blpage{2401}
(\byear{2024})
\end{barticle}
\endbibitem

\bibitem[\protect\citeauthoryear{Xu et~al.}{2024}]{xu2024resilience}
\begin{barticle}
\bauthor{\bsnm{Xu}, \binits{L.}},
\bauthor{\bsnm{Feng}, \binits{K.}},
\bauthor{\bsnm{Lin}, \binits{N.}},
\bauthor{\bsnm{Perera}, \binits{A.}},
\bauthor{\bsnm{Poor}, \binits{H.V.}},
\bauthor{\bsnm{Xie}, \binits{L.}},
\bauthor{\bsnm{Ji}, \binits{C.}},
\bauthor{\bsnm{Sun}, \binits{X.A.}},
\bauthor{\bsnm{Guo}, \binits{Q.}},
\bauthor{\bsnm{O’Malley}, \binits{M.}}:
\batitle{Resilience of renewable power systems under climate risks}.
\bjtitle{Nature Reviews Electrical Engineering}
\bvolume{1}(\bissue{1}),
\bfpage{53}--\blpage{66}
(\byear{2024})
\end{barticle}
\endbibitem

\bibitem[\protect\citeauthoryear{Damiani et~al.}{2024}]{damiani2024exploring}
\begin{barticle}
\bauthor{\bsnm{Damiani}, \binits{A.}},
\bauthor{\bsnm{Ishizaki}, \binits{N.N.}},
\bauthor{\bsnm{Sasaki}, \binits{H.}},
\bauthor{\bsnm{Feron}, \binits{S.}},
\bauthor{\bsnm{Cordero}, \binits{R.R.}}:
\batitle{Exploring super-resolution spatial downscaling of several meteorological variables and potential applications for photovoltaic power}.
\bjtitle{Scientific Reports}
\bvolume{14}(\bissue{1}),
\bfpage{7254}
(\byear{2024})
\end{barticle}
\endbibitem

\bibitem[\protect\citeauthoryear{Buster et~al.}{2024}]{Buster2024HighresolutionMW}
\begin{botherref}
\oauthor{\bsnm{Buster}, \binits{G.}},
\oauthor{\bsnm{Benton}, \binits{B.N.}},
\oauthor{\bsnm{Glaws}, \binits{A.}},
\oauthor{\bsnm{King}, \binits{R.N.}}:
High-resolution meteorology with climate change impacts from global climate model data using generative machine learning.
Nature Energy
(2024)
\end{botherref}
\endbibitem

\bibitem[\protect\citeauthoryear{Craig et~al.}{2022}]{craig2022overcoming}
\begin{barticle}
\bauthor{\bsnm{Craig}, \binits{M.T.}},
\bauthor{\bsnm{Wohland}, \binits{J.}},
\bauthor{\bsnm{Stoop}, \binits{L.P.}},
\bauthor{\bsnm{Kies}, \binits{A.}},
\bauthor{\bsnm{Pickering}, \binits{B.}},
\bauthor{\bsnm{Bloomfield}, \binits{H.C.}},
\bauthor{\bsnm{Browell}, \binits{J.}},
\bauthor{\bsnm{De~Felice}, \binits{M.}},
\bauthor{\bsnm{Dent}, \binits{C.J.}},
\bauthor{\bsnm{Deroubaix}, \binits{A.}}, \betal:
\batitle{Overcoming the disconnect between energy system and climate modeling}.
\bjtitle{Joule}
\bvolume{6}(\bissue{7}),
\bfpage{1405}--\blpage{1417}
(\byear{2022})
\end{barticle}
\endbibitem

\bibitem[\protect\citeauthoryear{Laprise}{2008}]{laprise2008regional}
\begin{barticle}
\bauthor{\bsnm{Laprise}, \binits{R.}}:
\batitle{Regional climate modelling}.
\bjtitle{Journal of computational physics}
\bvolume{227}(\bissue{7}),
\bfpage{3641}--\blpage{3666}
(\byear{2008})
\end{barticle}
\endbibitem

\bibitem[\protect\citeauthoryear{Jerez et~al.}{2020}]{jerez2020spin}
\begin{barticle}
\bauthor{\bsnm{Jerez}, \binits{S.}},
\bauthor{\bsnm{L{\'o}pez-Romero}, \binits{J.M.}},
\bauthor{\bsnm{Turco}, \binits{M.}},
\bauthor{\bsnm{Lorente-Plazas}, \binits{R.}},
\bauthor{\bsnm{G{\'o}mez-Navarro}, \binits{J.J.}},
\bauthor{\bsnm{Jim{\'e}nez-Guerrero}, \binits{P.}},
\bauthor{\bsnm{Mont{\'a}vez}, \binits{J.P.}}:
\batitle{On the spin-up period in wrf simulations over europe: Trade-offs between length and seasonality}.
\bjtitle{Journal of Advances in Modeling Earth Systems}
\bvolume{12}(\bissue{4}),
\bfpage{2019}--\blpage{001945}
(\byear{2020})
\end{barticle}
\endbibitem

\bibitem[\protect\citeauthoryear{Short and Petch}{2022}]{short2022reducing}
\begin{barticle}
\bauthor{\bsnm{Short}, \binits{C.J.}},
\bauthor{\bsnm{Petch}, \binits{J.}}:
\batitle{Reducing the spin-up of a regional nwp system without data assimilation}.
\bjtitle{Quarterly Journal of the Royal Meteorological Society}
\bvolume{148}(\bissue{745}),
\bfpage{1623}--\blpage{1643}
(\byear{2022})
\end{barticle}
\endbibitem

\bibitem[\protect\citeauthoryear{Lo et~al.}{2008}]{lo2008assessment}
\begin{botherref}
\oauthor{\bsnm{Lo}, \binits{J.C.-F.}},
\oauthor{\bsnm{Yang}, \binits{Z.-L.}},
\oauthor{\bsnm{Pielke~Sr}, \binits{R.A.}}:
Assessment of three dynamical climate downscaling methods using the weather research and forecasting (wrf) model.
Journal of Geophysical Research: Atmospheres
\textbf{113}(D9)
(2008)
\end{botherref}
\endbibitem

\bibitem[\protect\citeauthoryear{Stevens et~al.}{2019}]{stevens2019dyamond}
\begin{barticle}
\bauthor{\bsnm{Stevens}, \binits{B.}},
\bauthor{\bsnm{Satoh}, \binits{M.}},
\bauthor{\bsnm{Auger}, \binits{L.}},
\bauthor{\bsnm{Biercamp}, \binits{J.}},
\bauthor{\bsnm{Bretherton}, \binits{C.S.}},
\bauthor{\bsnm{Chen}, \binits{X.}},
\bauthor{\bsnm{D{\"u}ben}, \binits{P.}},
\bauthor{\bsnm{Judt}, \binits{F.}},
\bauthor{\bsnm{Khairoutdinov}, \binits{M.}},
\bauthor{\bsnm{Klocke}, \binits{D.}}, \betal:
\batitle{Dyamond: the dynamics of the atmospheric general circulation modeled on non-hydrostatic domains}.
\bjtitle{Progress in Earth and Planetary Science}
\bvolume{6}(\bissue{1}),
\bfpage{1}--\blpage{17}
(\byear{2019})
\end{barticle}
\endbibitem

\bibitem[\protect\citeauthoryear{Stensrud}{2007}]{stensrud2007parameterization}
\begin{bbook}
\bauthor{\bsnm{Stensrud}, \binits{D.J.}}:
\bbtitle{Parameterization Schemes: Keys to Understanding Numerical Weather Prediction Models}.
\bpublisher{Cambridge University Press}, \blocation{???}
(\byear{2007})
\end{bbook}
\endbibitem

\bibitem[\protect\citeauthoryear{Wedi}{2014}]{wedi2014increasing}
\begin{barticle}
\bauthor{\bsnm{Wedi}, \binits{N.P.}}:
\batitle{Increasing horizontal resolution in numerical weather prediction and climate simulations: illusion or panacea?}
\bjtitle{Philosophical Transactions of the Royal Society A: Mathematical, Physical and Engineering Sciences}
\bvolume{372}(\bissue{2018}),
\bfpage{20130289}
(\byear{2014})
\end{barticle}
\endbibitem

\bibitem[\protect\citeauthoryear{Dujardin and Lehning}{2022}]{dujardin2022wind}
\begin{barticle}
\bauthor{\bsnm{Dujardin}, \binits{J.}},
\bauthor{\bsnm{Lehning}, \binits{M.}}:
\batitle{Wind-topo: Downscaling near-surface wind fields to high-resolution topography in highly complex terrain with deep learning}.
\bjtitle{Quarterly Journal of the Royal Meteorological Society}
\bvolume{148}(\bissue{744}),
\bfpage{1368}--\blpage{1388}
(\byear{2022})
\end{barticle}
\endbibitem

\bibitem[\protect\citeauthoryear{Miyamoto et~al.}{2013}]{miyamoto2013deep}
\begin{barticle}
\bauthor{\bsnm{Miyamoto}, \binits{Y.}},
\bauthor{\bsnm{Kajikawa}, \binits{Y.}},
\bauthor{\bsnm{Yoshida}, \binits{R.}},
\bauthor{\bsnm{Yamaura}, \binits{T.}},
\bauthor{\bsnm{Yashiro}, \binits{H.}},
\bauthor{\bsnm{Tomita}, \binits{H.}}:
\batitle{Deep moist atmospheric convection in a subkilometer global simulation}.
\bjtitle{Geophysical Research Letters}
\bvolume{40}(\bissue{18}),
\bfpage{4922}--\blpage{4926}
(\byear{2013})
\end{barticle}
\endbibitem

\bibitem[\protect\citeauthoryear{Rasp et~al.}{2018}]{rasp2018deep}
\begin{barticle}
\bauthor{\bsnm{Rasp}, \binits{S.}},
\bauthor{\bsnm{Pritchard}, \binits{M.S.}},
\bauthor{\bsnm{Gentine}, \binits{P.}}:
\batitle{Deep learning to represent subgrid processes in climate models}.
\bjtitle{Proceedings of the national academy of sciences}
\bvolume{115}(\bissue{39}),
\bfpage{9684}--\blpage{9689}
(\byear{2018})
\end{barticle}
\endbibitem

\bibitem[\protect\citeauthoryear{Leutbecher and Palmer}{2008}]{leutbecher2008ensemble}
\begin{barticle}
\bauthor{\bsnm{Leutbecher}, \binits{M.}},
\bauthor{\bsnm{Palmer}, \binits{T.N.}}:
\batitle{Ensemble forecasting}.
\bjtitle{Journal of computational physics}
\bvolume{227}(\bissue{7}),
\bfpage{3515}--\blpage{3539}
(\byear{2008})
\end{barticle}
\endbibitem

\bibitem[\protect\citeauthoryear{Zhu et~al.}{2002}]{zhu2002economic}
\begin{barticle}
\bauthor{\bsnm{Zhu}, \binits{Y.}},
\bauthor{\bsnm{Toth}, \binits{Z.}},
\bauthor{\bsnm{Wobus}, \binits{R.}},
\bauthor{\bsnm{Richardson}, \binits{D.}},
\bauthor{\bsnm{Mylne}, \binits{K.}}:
\batitle{The economic value of ensemble-based weather forecasts}.
\bjtitle{Bulletin of the American Meteorological Society}
\bvolume{83}(\bissue{1}),
\bfpage{73}--\blpage{84}
(\byear{2002})
\end{barticle}
\endbibitem

\bibitem[\protect\citeauthoryear{Satoh et~al.}{2019}]{satoh2019global}
\begin{barticle}
\bauthor{\bsnm{Satoh}, \binits{M.}},
\bauthor{\bsnm{Stevens}, \binits{B.}},
\bauthor{\bsnm{Judt}, \binits{F.}},
\bauthor{\bsnm{Khairoutdinov}, \binits{M.}},
\bauthor{\bsnm{Lin}, \binits{S.-J.}},
\bauthor{\bsnm{Putman}, \binits{W.M.}},
\bauthor{\bsnm{D{\"u}ben}, \binits{P.}}:
\batitle{Global cloud-resolving models}.
\bjtitle{Current Climate Change Reports}
\bvolume{5},
\bfpage{172}--\blpage{184}
(\byear{2019})
\end{barticle}
\endbibitem

\bibitem[\protect\citeauthoryear{Buster et~al.}{2024}]{buster2024high}
\begin{botherref}
\oauthor{\bsnm{Buster}, \binits{G.}},
\oauthor{\bsnm{Benton}, \binits{B.N.}},
\oauthor{\bsnm{Glaws}, \binits{A.}},
\oauthor{\bsnm{King}, \binits{R.N.}}:
High-resolution meteorology with climate change impacts from global climate model data using generative machine learning.
Nature Energy,
1--13
(2024)
\end{botherref}
\endbibitem

\bibitem[\protect\citeauthoryear{Bronstein et~al.}{2017}]{bronstein2017geometric}
\begin{barticle}
\bauthor{\bsnm{Bronstein}, \binits{M.M.}},
\bauthor{\bsnm{Bruna}, \binits{J.}},
\bauthor{\bsnm{LeCun}, \binits{Y.}},
\bauthor{\bsnm{Szlam}, \binits{A.}},
\bauthor{\bsnm{Vandergheynst}, \binits{P.}}:
\batitle{Geometric deep learning: going beyond euclidean data}.
\bjtitle{IEEE Signal Processing Magazine}
\bvolume{34}(\bissue{4}),
\bfpage{18}--\blpage{42}
(\byear{2017})
\end{barticle}
\endbibitem

\bibitem[\protect\citeauthoryear{Pan et~al.}{2019}]{pan2019improving}
\begin{barticle}
\bauthor{\bsnm{Pan}, \binits{B.}},
\bauthor{\bsnm{Hsu}, \binits{K.}},
\bauthor{\bsnm{AghaKouchak}, \binits{A.}},
\bauthor{\bsnm{Sorooshian}, \binits{S.}}:
\batitle{Improving precipitation estimation using convolutional neural network}.
\bjtitle{Water Resources Research}
\bvolume{55}(\bissue{3}),
\bfpage{2301}--\blpage{2321}
(\byear{2019})
\end{barticle}
\endbibitem

\bibitem[\protect\citeauthoryear{Gilpin}{2023}]{PhysRevResearch.5.043252}
\begin{barticle}
\bauthor{\bsnm{Gilpin}, \binits{W.}}:
\batitle{Model scale versus domain knowledge in statistical forecasting of chaotic systems}.
\bjtitle{Phys. Rev. Res.}
\bvolume{5},
\bfpage{043252}
(\byear{2023})
\doiurl{10.1103/PhysRevResearch.5.043252}
\end{barticle}
\endbibitem

\bibitem[\protect\citeauthoryear{Pan et~al.}{2021}]{pan2021learning}
\begin{barticle}
\bauthor{\bsnm{Pan}, \binits{B.}},
\bauthor{\bsnm{Anderson}, \binits{G.J.}},
\bauthor{\bsnm{Goncalves}, \binits{A.}},
\bauthor{\bsnm{Lucas}, \binits{D.D.}},
\bauthor{\bsnm{Bonfils}, \binits{C.J.}},
\bauthor{\bsnm{Lee}, \binits{J.}},
\bauthor{\bsnm{Tian}, \binits{Y.}},
\bauthor{\bsnm{Ma}, \binits{H.-Y.}}:
\batitle{Learning to correct climate projection biases}.
\bjtitle{Journal of Advances in Modeling Earth Systems}
\bvolume{13}(\bissue{10}),
\bfpage{2021}--\blpage{002509}
(\byear{2021})
\end{barticle}
\endbibitem

\bibitem[\protect\citeauthoryear{Israeli and Goldenfeld}{2004}]{wolfram2002computational}
\begin{barticle}
\bauthor{\bsnm{Israeli}, \binits{N.}},
\bauthor{\bsnm{Goldenfeld}, \binits{N.}}:
\batitle{Computational irreducibility and the predictability of complex physical systems}.
\bjtitle{Phys. Rev. Lett.}
\bvolume{92},
\bfpage{074105}
(\byear{2004})
\doiurl{10.1103/PhysRevLett.92.074105}
\end{barticle}
\endbibitem

\bibitem[\protect\citeauthoryear{Ho et~al.}{2020}]{ho2020denoising}
\begin{barticle}
\bauthor{\bsnm{Ho}, \binits{J.}},
\bauthor{\bsnm{Jain}, \binits{A.}},
\bauthor{\bsnm{Abbeel}, \binits{P.}}:
\batitle{Denoising diffusion probabilistic models}.
\bjtitle{Advances in neural information processing systems}
\bvolume{33},
\bfpage{6840}--\blpage{6851}
(\byear{2020})
\end{barticle}
\endbibitem

\bibitem[\protect\citeauthoryear{Song et~al.}{2020}]{song2020score}
\begin{botherref}
\oauthor{\bsnm{Song}, \binits{Y.}},
\oauthor{\bsnm{Sohl-Dickstein}, \binits{J.}},
\oauthor{\bsnm{Kingma}, \binits{D.P.}},
\oauthor{\bsnm{Kumar}, \binits{A.}},
\oauthor{\bsnm{Ermon}, \binits{S.}},
\oauthor{\bsnm{Poole}, \binits{B.}}:
Score-based generative modeling through stochastic differential equations.
arXiv preprint arXiv:2011.13456
(2020)
\end{botherref}
\endbibitem

\bibitem[\protect\citeauthoryear{Song et~al.}{2021}]{song2021solving}
\begin{botherref}
\oauthor{\bsnm{Song}, \binits{Y.}},
\oauthor{\bsnm{Shen}, \binits{L.}},
\oauthor{\bsnm{Xing}, \binits{L.}},
\oauthor{\bsnm{Ermon}, \binits{S.}}:
Solving inverse problems in medical imaging with score-based generative models.
arXiv preprint arXiv:2111.08005
(2021)
\end{botherref}
\endbibitem

\bibitem[\protect\citeauthoryear{Feng et~al.}{2023}]{feng2023score}
\begin{bchapter}
\bauthor{\bsnm{Feng}, \binits{B.T.}},
\bauthor{\bsnm{Smith}, \binits{J.}},
\bauthor{\bsnm{Rubinstein}, \binits{M.}},
\bauthor{\bsnm{Chang}, \binits{H.}},
\bauthor{\bsnm{Bouman}, \binits{K.L.}},
\bauthor{\bsnm{Freeman}, \binits{W.T.}}:
\bctitle{Score-based diffusion models as principled priors for inverse imaging}.
In: \bbtitle{Proceedings of the IEEE/CVF International Conference on Computer Vision},
pp. \bfpage{10520}--\blpage{10531}
(\byear{2023})
\end{bchapter}
\endbibitem

\bibitem[\protect\citeauthoryear{Bouman and Buzzard}{2023}]{bouman2023generative}
\begin{bchapter}
\bauthor{\bsnm{Bouman}, \binits{C.A.}},
\bauthor{\bsnm{Buzzard}, \binits{G.T.}}:
\bctitle{Generative plug and play: Posterior sampling for inverse problems}.
In: \bbtitle{2023 59th Annual Allerton Conference on Communication, Control, and Computing (Allerton)},
pp. \bfpage{1}--\blpage{7}
(\byear{2023}).
\bcomment{IEEE}
\end{bchapter}
\endbibitem

\bibitem[\protect\citeauthoryear{Habring and Holler}{2024}]{habring2024neural}
\begin{botherref}
\oauthor{\bsnm{Habring}, \binits{A.}},
\oauthor{\bsnm{Holler}, \binits{M.}}:
Neural-network-based regularization methods for inverse problems in imaging.
GAMM-Mitteilungen,
202470004
(2024)
\end{botherref}
\endbibitem

\bibitem[\protect\citeauthoryear{Page et~al.}{2021}]{page2021revealing}
\begin{barticle}
\bauthor{\bsnm{Page}, \binits{J.}},
\bauthor{\bsnm{Brenner}, \binits{M.P.}},
\bauthor{\bsnm{Kerswell}, \binits{R.R.}}:
\batitle{Revealing the state space of turbulence using machine learning}.
\bjtitle{Physical Review Fluids}
\bvolume{6}(\bissue{3}),
\bfpage{034402}
(\byear{2021})
\end{barticle}
\endbibitem

\bibitem[\protect\citeauthoryear{Venkatakrishnan et~al.}{2013}]{6737048}
\begin{bchapter}
\bauthor{\bsnm{Venkatakrishnan}, \binits{S.V.}},
\bauthor{\bsnm{Bouman}, \binits{C.A.}},
\bauthor{\bsnm{Wohlberg}, \binits{B.}}:
\bctitle{Plug-and-play priors for model based reconstruction}.
In: \bbtitle{2013 IEEE Global Conference on Signal and Information Processing},
pp. \bfpage{945}--\blpage{948}
(\byear{2013}).
\doiurl{10.1109/GlobalSIP.2013.6737048}
\end{bchapter}
\endbibitem

\bibitem[\protect\citeauthoryear{Pan et~al.}{2023}]{pan2023probabilistic}
\begin{botherref}
\oauthor{\bsnm{Pan}, \binits{B.}},
\oauthor{\bsnm{Wang}, \binits{L.-Y.}},
\oauthor{\bsnm{Zhang}, \binits{F.}},
\oauthor{\bsnm{Duan}, \binits{Q.}},
\oauthor{\bsnm{Li}, \binits{X.}},
\oauthor{\bsnm{Pan}, \binits{X.}},
\oauthor{\bsnm{Chen}, \binits{X.}},
\oauthor{\bsnm{Ling}, \binits{F.}},
\oauthor{\bsnm{Wang}, \binits{S.}},
\oauthor{\bsnm{Pan}, \binits{M.}}, et al.:
Probabilistic diffusion model for stochastic parameterization--a case example of numerical precipitation estimation.
Authorea Preprints
(2023)
\end{botherref}
\endbibitem

\bibitem[\protect\citeauthoryear{Skamarock et~al.}{2008}]{skamarock2008description}
\begin{barticle}
\bauthor{\bsnm{Skamarock}, \binits{W.C.}},
\bauthor{\bsnm{Klemp}, \binits{J.B.}},
\bauthor{\bsnm{Dudhia}, \binits{J.}},
\bauthor{\bsnm{Gill}, \binits{D.O.}},
\bauthor{\bsnm{Barker}, \binits{D.M.}},
\bauthor{\bsnm{Duda}, \binits{M.G.}},
\bauthor{\bsnm{Huang}, \binits{X.-Y.}},
\bauthor{\bsnm{Wang}, \binits{W.}},
\bauthor{\bsnm{Powers}, \binits{J.G.}}, \betal:
\batitle{A description of the advanced research wrf version 3}.
\bjtitle{NCAR technical note}
\bvolume{475}(\bissue{125}),
\bfpage{10}--\blpage{5065}
(\byear{2008})
\end{barticle}
\endbibitem

\bibitem[\protect\citeauthoryear{Hersbach et~al.}{2020}]{hersbach2020era5}
\begin{barticle}
\bauthor{\bsnm{Hersbach}, \binits{H.}},
\bauthor{\bsnm{Bell}, \binits{B.}},
\bauthor{\bsnm{Berrisford}, \binits{P.}},
\bauthor{\bsnm{Hirahara}, \binits{S.}},
\bauthor{\bsnm{Hor{\'a}nyi}, \binits{A.}},
\bauthor{\bsnm{Mu{\~n}oz-Sabater}, \binits{J.}},
\bauthor{\bsnm{Nicolas}, \binits{J.}},
\bauthor{\bsnm{Peubey}, \binits{C.}},
\bauthor{\bsnm{Radu}, \binits{R.}},
\bauthor{\bsnm{Schepers}, \binits{D.}}, \betal:
\batitle{The era5 global reanalysis}.
\bjtitle{Quarterly Journal of the Royal Meteorological Society}
\bvolume{146}(\bissue{730}),
\bfpage{1999}--\blpage{2049}
(\byear{2020})
\end{barticle}
\endbibitem

\bibitem[\protect\citeauthoryear{Chao et~al.}{2025}]{chaoj2025}
\begin{barticle}
\bauthor{\bsnm{Chao}, \binits{J.}},
\bauthor{\bsnm{Pan}, \binits{B.}},
\bauthor{\bsnm{Chen}, \binits{Q.}},
\bauthor{\bsnm{Yang}, \binits{S.}},
\bauthor{\bsnm{Wang}, \binits{J.}},
\bauthor{\bsnm{Nai}, \binits{C.}},
\bauthor{\bsnm{Zheng}, \binits{Y.}},
\bauthor{\bsnm{Li}, \binits{X.}},
\bauthor{\bsnm{Yuan}, \binits{H.}},
\bauthor{\bsnm{Chen}, \binits{X.}},
\bauthor{\bsnm{Lu}, \binits{B.}},
\bauthor{\bsnm{Xiao}, \binits{Z.}}:
\batitle{Learning to infer weather states using partial observations}.
\bjtitle{Journal of Geophysical Research: Machine Learning and Computation}
\bvolume{2}(\bissue{1}),
\bfpage{2024}--\blpage{000260}
(\byear{2025})
\doiurl{10.1029/2024JH000260}
\end{barticle}
\endbibitem

\bibitem[\protect\citeauthoryear{Yang et~al.}{2025}]{pan2025GAP}
\begin{botherref}
\oauthor{\bsnm{Yang}, \binits{S.}},
\oauthor{\bsnm{Nai}, \binits{C.}},
\oauthor{\bsnm{Liu}, \binits{X.}},
\oauthor{\bsnm{Li}, \binits{W.}},
\oauthor{\bsnm{Chao}, \binits{J.}},
\oauthor{\bsnm{Wang}, \binits{J.}},
\oauthor{\bsnm{Wang}, \binits{L.}},
\oauthor{\bsnm{Li}, \binits{X.}},
\oauthor{\bsnm{Chen}, \binits{X.}},
\oauthor{\bsnm{Lu}, \binits{B.}},
\oauthor{\bsnm{Xiao}, \binits{Z.}},
\oauthor{\bsnm{Boers}, \binits{N.}},
\oauthor{\bsnm{Yuan}, \binits{H.}},
\oauthor{\bsnm{Pan}, \binits{B.}}:
Generative assimilation and prediction for weather and climate.
arXiv preprint arXiv:2503.03038
(2025)
\end{botherref}
\endbibitem

\bibitem[\protect\citeauthoryear{Wang et~al.}{2021}]{wang2021review}
\begin{barticle}
\bauthor{\bsnm{Wang}, \binits{Y.}},
\bauthor{\bsnm{Zou}, \binits{R.}},
\bauthor{\bsnm{Liu}, \binits{F.}},
\bauthor{\bsnm{Zhang}, \binits{L.}},
\bauthor{\bsnm{Liu}, \binits{Q.}}:
\batitle{A review of wind speed and wind power forecasting with deep neural networks}.
\bjtitle{Applied Energy}
\bvolume{304},
\bfpage{117766}
(\byear{2021})
\end{barticle}
\endbibitem

\bibitem[\protect\citeauthoryear{Powers et~al.}{2017}]{powers2017weather}
\begin{barticle}
\bauthor{\bsnm{Powers}, \binits{J.G.}},
\bauthor{\bsnm{Klemp}, \binits{J.B.}},
\bauthor{\bsnm{Skamarock}, \binits{W.C.}},
\bauthor{\bsnm{Davis}, \binits{C.A.}},
\bauthor{\bsnm{Dudhia}, \binits{J.}},
\bauthor{\bsnm{Gill}, \binits{D.O.}},
\bauthor{\bsnm{Coen}, \binits{J.L.}},
\bauthor{\bsnm{Gochis}, \binits{D.J.}},
\bauthor{\bsnm{Ahmadov}, \binits{R.}},
\bauthor{\bsnm{Peckham}, \binits{S.E.}}, \betal:
\batitle{The weather research and forecasting model: Overview, system efforts, and future directions}.
\bjtitle{Bulletin of the American Meteorological Society}
\bvolume{98}(\bissue{8}),
\bfpage{1717}--\blpage{1737}
(\byear{2017})
\end{barticle}
\endbibitem

\bibitem[\protect\citeauthoryear{Meng et~al.}{2021}]{meng2021sdedit}
\begin{botherref}
\oauthor{\bsnm{Meng}, \binits{C.}},
\oauthor{\bsnm{He}, \binits{Y.}},
\oauthor{\bsnm{Song}, \binits{Y.}},
\oauthor{\bsnm{Song}, \binits{J.}},
\oauthor{\bsnm{Wu}, \binits{J.}},
\oauthor{\bsnm{Zhu}, \binits{J.-Y.}},
\oauthor{\bsnm{Ermon}, \binits{S.}}:
Sdedit: Guided image synthesis and editing with stochastic differential equations.
arXiv preprint arXiv:2108.01073
(2021)
\end{botherref}
\endbibitem

\bibitem[\protect\citeauthoryear{Ronneberger et~al.}{2015}]{ronneberger2015u}
\begin{bchapter}
\bauthor{\bsnm{Ronneberger}, \binits{O.}},
\bauthor{\bsnm{Fischer}, \binits{P.}},
\bauthor{\bsnm{Brox}, \binits{T.}}:
\bctitle{U-net: Convolutional networks for biomedical image segmentation}.
In: \bbtitle{Medical Image Computing and Computer-assisted intervention--MICCAI 2015: 18th International Conference, Munich, Germany, October 5-9, 2015, Proceedings, Part III 18},
pp. \bfpage{234}--\blpage{241}
(\byear{2015}).
\bcomment{Springer}
\end{bchapter}
\endbibitem

\bibitem[\protect\citeauthoryear{Kingma and Ba}{2014}]{kingma2014adam}
\begin{botherref}
\oauthor{\bsnm{Kingma}, \binits{D.P.}},
\oauthor{\bsnm{Ba}, \binits{J.}}:
Adam: A method for stochastic optimization.
arXiv preprint arXiv:1412.6980
(2014)
\end{botherref}
\endbibitem

\end{thebibliography}

\end{document}